\documentclass[10pt,twocolumn,letterpaper]{article}

\usepackage[pagenumbers]{wacv} 

\usepackage[accsupp]{axessibility}  

\usepackage{booktabs}
\usepackage[table]{xcolor}

\usepackage{graphicx}
\usepackage{amsmath}
\usepackage{amssymb}
\usepackage{booktabs}

\usepackage{amsthm}

\usepackage{algorithm}
\usepackage{algorithmic}

\usepackage{enumitem}

\usepackage[pagebackref,breaklinks,colorlinks]{hyperref}

\usepackage[capitalize]{cleveref}

\crefname{enumi}{Condition}{Conditions}
\crefname{section}{Sec.}{Secs.}
\Crefname{section}{Section}{Sections}
\Crefname{table}{Table}{Tables}
\crefname{table}{Tab.}{Tabs.}
\crefname{algorithm}{Alg.}{Algs.}

\newcommand\blfootnote[1]{%
  \begingroup
  \renewcommand\thefootnote{}\footnote{#1}%
  \addtocounter{footnote}{-1}%
  \endgroup
}

\providecommand{\lin}[1]{\ensuremath{\left\langle #1 \right\rangle}}

\providecommand{\norm}[1]{\ensuremath{\left\lVert#1\right\rVert}}

\providecommand{\R}{\mathbb{R}} 

\DeclareMathOperator*{\argmin}{arg\,min}
\DeclareMathOperator*{\argmax}{arg\,max}

\providecommand{\pp}{\mathbf{p}}

\providecommand{\tt}{\mathbf{t}}

\providecommand{\xx}{\mathbf{x}}

  \providecommand{\cB}{\mathcal{B}}
  \providecommand{\cC}{\mathcal{C}}
  \providecommand{\cD}{\mathcal{D}}

  \providecommand{\cL}{\mathcal{L}}
  
  \providecommand{\cN}{\mathcal{N}}
  
  \providecommand{\cP}{\mathcal{P}}

  \providecommand{\cT}{\mathcal{T}}

  \providecommand{\cX}{\mathcal{X}}

\newtheorem{claim}{Claim}

\usepackage{multirow}

\usepackage[acronym]{glossaries}
\newacronym{dml}{DML}{Deep Metric Learning}
\newacronym{sl}{SL}{SuperLoss}
\newacronym{ms}{MS}{Multi-similarity}
\newacronym{sop}{SOP}{Standard Online Products}
\newacronym{llm}{LLM}{Large Language Model}
\newacronym{plg}{PLG}{Pseudolabel Language Guidance}
\newacronym{cl}{CL}{Curriculum Learning}
\newacronym{nmi}{NMI}{Normalized Mutual Information}

\begin{document}

\title{ProcSim: Proxy-based Confidence for Robust Similarity Learning}

\author{
Oriol Barbany$^{1\dagger}$
\hspace{2ex} Xiaofan Lin$^{2}$
\hspace{2ex} Muhammet Bastan$^{2}$
\hspace{2ex} Arnab Dhua$^{2}$ \\ \vspace{1ex}
\normalsize{${}^{1}$Institut de Robòtica i Informàtica Industrial, CSIC-UPC} \hskip3ex \normalsize{${}^{2}$Visual Search \& AR, Amazon} \\[-1ex] {\tt\small obarbany@iri.upc.edu, \{xiaofanl,mbastan,adhua\}@amazon.com} \\[-1ex]
}
\maketitle

\iftoggle{wacvfinal}{\blfootnote{$\dagger$ Work performed during an internship at Amazon.}}{}

\begin{abstract}
\gls*{dml} methods aim at learning an embedding space in which distances are closely related to the inherent semantic similarity of the inputs.
Previous studies have shown that
popular benchmark datasets often contain numerous
wrong labels, and \gls*{dml} methods are susceptible to them.
Intending to study the effect of realistic noise, we create an ontology of the classes in a dataset and use it to simulate semantically coherent labeling mistakes. To train robust \gls*{dml} models, we propose ProcSim, a simple framework that assigns a confidence score to each sample using the normalized distance to its class representative.
The experimental results show that the proposed method achieves state-of-the-art performance on the \gls*{dml} benchmark datasets injected with uniform and the proposed semantically coherent noise.
\end{abstract}
\glsresetall

\vspace{-1em}

\section{Introduction}
\label{sec:intro}

The problem of quantifying the similarity between images is typically framed in the context of metric learning, which aims at learning a metric space in which distances closely relate to underlying semantic similarities.
\Gls*{dml} is based on transforming the images using a neural network and then applying a predefined metric, \eg, the Euclidean distance, or cosine similarity.

Identifying visual similarities is crucial for tasks such as image retrieval \cite{semantic_granularity}, zero-shot learning \cite{zero-shot}, and person identification \cite{triplet_loss,Shuai2022}. Solving these problems with \gls*{dml} allows the introduction of new classes without retraining, a desirable feature in applications such as retail \cite{retail_product_recognition}. Moreover, the learned similarity model can be easily paired with efficient nearest-neighbor inference techniques \cite{faiss}.

\gls*{dml} requires labeled datasets, but manual labeling is cumbersome and, in some cases, infeasible. Automated labeling, while efficient, introduces errors like duplicates and irrelevant images, often necessitating manual correction \cite{sop_dataset}. Conversely, manual annotations often involve non-expert annotators on crowdsourcing platforms, leading to occasional labeling errors \cite{cars196}. Labeling mistakes are especially problematic for \gls*{dml}, which suffer a higher drop in performance than classification models as the number of noisy labels increases \cite{dereka22}.

While \gls*{dml} with noisy labels has garnered attention, prior research has mostly focused on building robust models against uniform noise \cite{Yan2021AdaptiveHS,prism,zeng2022}. However, due to the annotation techniques in image retrieval, real datasets often exhibit noise concentrated in clusters of similar images \cite{dereka22}.

\begin{figure}[t]
    \centering
    \includegraphics[width=\columnwidth]{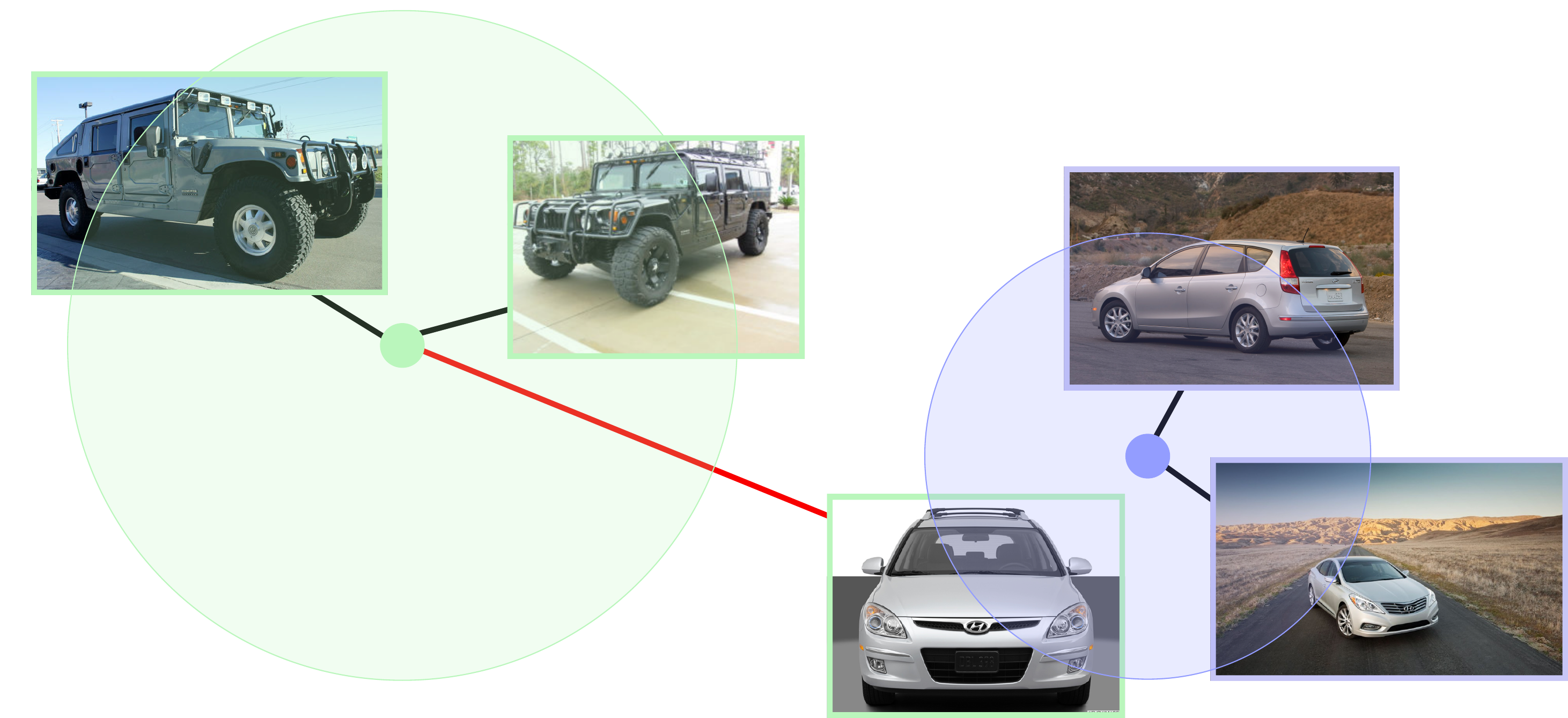}
    \vspace{-20pt}
    \caption{ProcSim handles incorrect labels by reducing the contribution of samples whose learned embeddings are too far away from their class representatives.}
    \label{fig:first_figure}
    \vspace{-18pt}
\end{figure}

This paper proposes ProcSim, a new confidence-aware framework for training robust \gls*{dml} models by estimating the reliability of samples in an unsupervised fashion. To test the benefits of our method on noisy datasets, we present a new procedure for injecting semantically coherent label errors.
The empirical results show the superior performance of ProcSim trained on benchmark datasets injected with uniform and the proposed semantic noise in front of alternative approaches.

The main contributions of this paper are:
\vspace{-5pt}
\begin{itemize}
    \item We propose ProcSim, a novel framework for robust visual similarity learning usable on top of any general-purpose \gls*{dml} loss to improve performance on noisy datasets. ProcSim assigns a per-sample confidence that indicates the reliability of its label and is used to determine the influence of such a sample during training.
    \item We introduce a new noise model based on swapping semantically similar class labels. \cref{sec:semantic_noise_generation} describes how to automatically obtain a hierarchy of the classes in a dataset and use it to inject label noise.
\end{itemize}

\section{Related work}
\label{sec:related_work}

\subsection{Learning with noisy labels}

Some approaches dealing with noisy data estimate the noise transition matrix \cite{patrini2017making,xia2019,zeng2022}, which requires prior knowledge or a subset of clean data. Another class of methods uses the model predictions to correct the labels \cite{li2020dividemix,unicon,zeng2022}. However, this technique can lead to confirmation bias, where prediction errors accumulate and harm performance \cite{Yan_2022_CVPR}. Alternatively, one can estimate which samples are incorrectly annotated \cite{han2018co,li2020dividemix,ibrahimi22}. These methods typically assume that significant loss instances can be associated with incorrect labels, a technique commonly known as the small-loss trick.

The small-loss trick is rooted in the observation that deep neural networks often learn clean samples before noisy samples \cite{arpit_memorization}, resulting in inputs with accurate labels exhibiting lower-magnitude losses \cite{chen2019understanding}.

Some works on noisy classification train two semi-independent networks that exchange information about noisy samples to prevent their memorization \cite{han2018co,lee2017cleannet,yu2019does,wang2019}. Directly adopting these methods to \gls*{dml} is not feasible \cite{Yan2021AdaptiveHS}, but there exist similar approaches in the \gls*{dml} literature using self-distillation to determine soft labels \cite{zeng2022} or detect noisy samples \cite{ibrahimi22}.

If the noise probability is known and the small-loss trick assumption is satisfied, one can spot noisy samples as those whose loss value is over a given percentile determined by the noise probability \cite{prism,ibrahimi22}. However, the amount of noise present in a dataset is generally unknown.

Under the more realistic case where the noise probability is unknown, an interesting approach is to fit a bimodal distribution to explain the loss values \cite{li2020dividemix}. Then, following the small-loss trick, the samples belonging to the distribution with the higher mode are treated as noisy.

Once noisy samples are detected, we can split the training dataset into disjoint sets representing correct and incorrect labels. In the context of \gls*{dml}, when we identify a sample as noisy, we can discard it \cite{prism} or only consider it for negative interactions \cite{ibrahimi22}.

Instead of treating all correct and incorrect samples equally, an option is to use a confidence-aware loss, in which the loss amplitude is modulated proportionally to the sample confidence \cite{ssl_confidence}. Ideally, noisy samples will be assigned a low confidence score to reduce or even suppress their contribution. SuperLoss \cite{superloss} offers a task-agnostic approach to converting any loss into a confident-aware loss without additional learnable parameters.

\subsection{Inter-class similarities}

Inter-class similarities can be considered by clustering image features and creating a class tree \cite{hierarchical_triplet_loss} or promoting the clusters formed during training \cite{hierarchical_proxy}. Another approach is to modify a margin-based objective so that the margin depends on the attribute similarity \cite{semantic_granularity}.
One compelling alternative is to distill the knowledge of a \gls*{llm} to learn semantically consistent metric spaces \cite{language_guidance}. One can also learn a hyperbolic space \cite{Ermolov_2022_CVPR,Yan2021AdaptiveHS}, which naturally embeds hierarchies.

\subsection{Non-uniform noise generation}

Swapping labels using semantic similarities results in plausible labeling mistakes and noisy samples that are more challenging to spot \cite{li2020dividemix}. Using this idea, some works on noisy classification considered injecting class label errors based on the structure of recurring mistakes in real datasets, \eg, Truck$\rightarrow$Automobile, Bird$\rightarrow$Airplane, and Dog$\leftrightarrow$ Cat \cite{li2020dividemix,unicon}. However, inferring these rules is specific to each dataset and requires statistics about the errors.

In the context of noisy \gls*{dml}, Liu \etal \cite{prism} proposed an iterative procedure to introduce noise. In each iteration, they choose a class and group its samples by employing a similarity measure computed using a pre-trained \gls*{dml} model. Then, they assign the same class label to all cluster members. Although this method incorporates a notion of visual similarity for the clustering step, label assignment is performed uniformly at random, and the number of classes decreases at each iteration. Dereka \etal \cite{dereka22} introduced the large and small class label noise models based on only corrupting the most frequent or rarest classes. While this method restricts the set of possible labels assigned (asymmetric noise), the choice is purely based on class frequencies, not semantics.

\section{Methodology}
\label{sec:methodology}

\subsection{Preliminaries}

Let $\cD:=\{(\xx_i,y_i)\}_{i\in[n]}$ be a dataset with pairs of images $\xx_i \in \cX$ and class labels $y_i\in [C]$. \Gls*{dml} aims to learn a metric space $(\Psi, d)$ with fixed $d:\Psi\times \Psi \to \R$ and a learned transformation $\phi: \cX \to \Psi$ such that $d(\phi(\xx_i), \phi(\xx_j))<d(\phi(\xx_i), \phi(\xx_k))$ if $\xx_i$ is semantically more similar to $\xx_j$ than it is to $\xx_k$ \cite{metric_learning_book}. Commonly, the space $\Psi$ is normalized to the unit hypersphere for training stability \cite{triplet_loss,margin,divide_and_conquer}, and $d$ is chosen to be the Euclidean or cosine distance.

Instead of computing the confidence of the sample using a learnable model \cite{ssl_confidence,confidence_from_net_2,confidence_from_net_3}, we prefer to follow a parsimonious approach inspired by SuperLoss \cite{superloss}, a technique that computes a confidence score from the training loss and uses it for the task of automatic curriculum learning. In the curriculum learning training, the samples are fed in increasing order of difficulty, which improves the speed of convergence and the quality of the models obtained \cite{curriculum_learning,power_of_cl}.

For the \gls*{dml} problem, SuperLoss assigns a confidence $\sigma_{ij}$ to each pair of samples. Doing that requires an objective expressed as a double sum over pairs, \eg, the contrastive loss \cite{chopra2005learning}. For a pair of samples $(i,j)$ with loss $\ell_{ij}$, instead of directly minimizing $\mathbb{E}_{(i,j)}[\ell_{ij}]$ as in regular training, SuperLoss proposes to minimize
\begin{align}
\mathbb{E}_{(i,j)}\left[	\min_{\sigma_{ij}} (\ell_{ij} - {\tau_{ij}}) \sigma_{ij} + \lambda (\log \sigma_{ij})^2\right] \,,
	\label{eq:superloss_objective}
\end{align}
where $\lambda \in\R^+$, and $\tau_{ij}$ is the global average of all positive (resp.~negative) pair losses across all iterations if $y_i=y_j$  (resp. $y_i\neq y_j$). The optimization of the pair confidence has the closed form solution
\begin{align}
	\sigma_{ij} = \exp\left[{-W\left(\frac{1}{2}\max\left\lbrace -\frac{2}{e}, \frac{\ell_{ij} - {\tau_{ij}}}{\lambda}  \right\rbrace \right)}\right]\,,
	\label{eq:optimal_sigma_pair}
\end{align}
where $W(\cdot)$ is the principal branch of the Lambert W function. The authors of SuperLoss \cite{superloss} use this analytical solution to compute the optimal confidence and avoid the minimization in \cref{eq:superloss_objective}. The confidence is treated as a constant, meaning that they don't propagate gradients through it.

\begin{figure}[t!]
    \centering
    \includegraphics[width=\columnwidth]{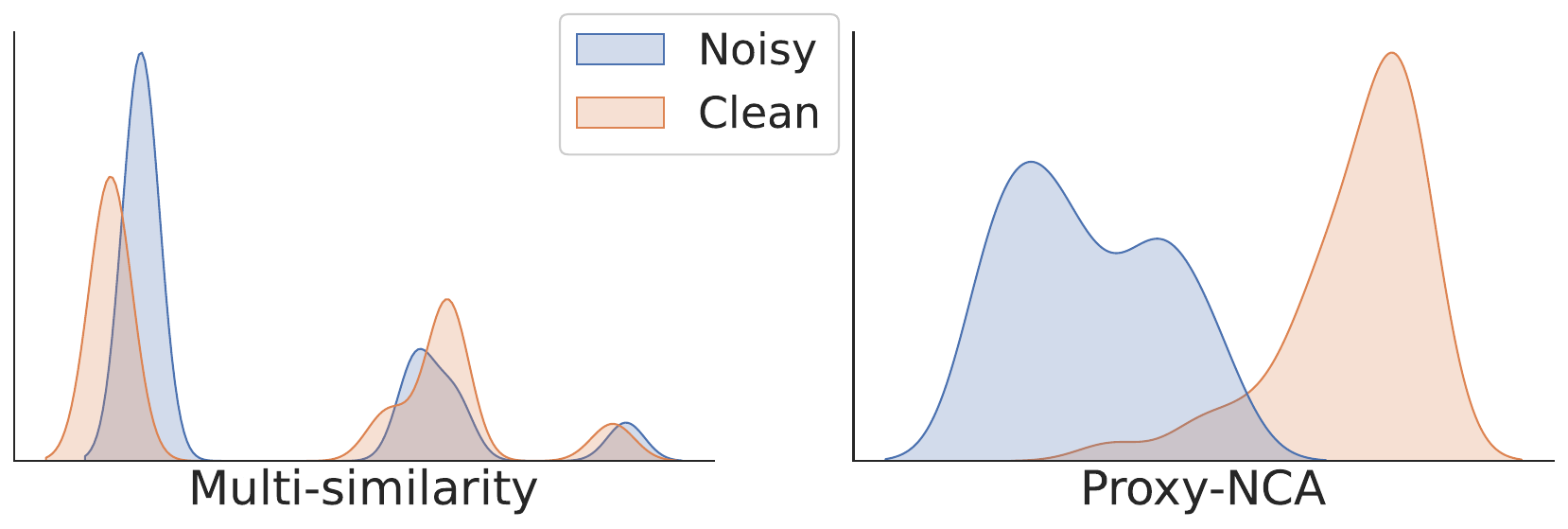}
    \caption{Distribution of loss values for clean and noisy samples in the late stages of training on CUB200 \cite{CUB_200_2011} with 50\% uniform noise. While the \gls*{ms} loss \cite{multisimilarity} is a powerful objective for training \gls*{dml} models, it is unsuited for label noise identification. Classification of noisy samples using Otsu's threshold \cite{otsu} achieved 50\% and 90\% recall, respectively. More details in the supplementary.
    }
    \label{fig:small_loss_trick}
    \vspace{-15pt}
\end{figure}

\subsection{Identifying noisy samples}
\label{sec:identify_noisy_samples}

Curriculum learning down-weights the contribution of challenging samples, sometimes resulting in the omission of noisy samples \cite{jiang2018mentornet,Lyu2020Curriculum}. However, inputs considered hard in the curriculum learning context change across iterations while the number of incorrect annotations in a dataset remain the same. Particularly for \gls*{dml}, the loss is obtained by considering interactions--pairs, triplets, or tuples of a higher order--with the other samples in a batch. Hence, large loss values may be either because of a wrong label of the anchor sample or others included in the considered interactions. Therefore, data points that are hard to explain under the training objective are not necessarily those with an incorrect class label.

\cref{fig:small_loss_trick} shows the distribution of noisy and clean samples when using two well-known \gls*{dml} losses. The \gls*{ms} \cite{multisimilarity} objective penalizes the positive pairs with lower similarity and the negative pairs with higher similarity. Thus, a clean sample interacting with a noisy one will almost exclusively consider the latter, which will cause large loss values. Hence, this loss is unsuited for spotting noisy samples.

Let $\{\pp_i\}_{i\in [C]}$ be a set of points representing classes and $\xx$ an unlabeled sample. The nearest neighbor search on $\phi$ returns $\argmax_{i\in[C]}\lin{\phi(\xx), \pp_i}$. $\text{Softmax}$ is a smooth approximation of $\argmax$, and replacing it in the previous expression yields a stochastic nearest neighbor classifier. The Proxy-NCA \cite{proxynca} loss for sample $i$, which we will refer to as $\ell_i^{\text{Proxy}}$, is precisely the negative $log(\cdot)$ of the probability that a stochastic nearest neighbor classifier assigns a sample to its correct label when $\{\pp_i\}_{i\in [C]}$ are class proxies. 

The class proxies are learnable embeddings representing data groups and have the desirable feature that they are robust to noisy labels \cite{kim2020proxy}. Therefore, even when some class contains wrong annotations, their proxies will be close to the embeddings of the clean samples of that class. Overall, Proxy-NCA loss is fundamentally a normalized distance to the class representative. This observation provides a theoretical explanation of why large sample loss values can be associated with a possibly incorrect label.

\subsection{Separating noisy and clean samples}

In \cref{fig:small_loss_trick}, we present some empirical evidence of the identifiability of noisy samples under the Proxy-NCA \cite{proxynca}. 
Indeed, the distribution follows a bimodal pattern, with wrongly annotated data points falling within the mode exhibiting higher losses. One option to separate clean and noisy samples is to use a Gaussian mixture model \cite{li2020dividemix}. However, this method assumes that each distribution is a Gaussian, which is not the case for the skewed distributions of clean and noisy samples in \cref{fig:small_loss_trick}. Moreover, this approach requires an iterative procedure to estimate the sufficient statistics of each distribution.

An alternative is using Otsu's method, a one-dimensional discrete analog of Fisher's discriminant analysis. This approach selects a threshold that minimizes the intra-class variance (equivalently, maximizing the inter-class variance) and is typically used to perform image thresholding. Otsu's method does not require any optimization, has no hyper-parameters, and achieves the same result as globally optimal $K$-means\cite{otsu_kmeans}. 

In \cref{alg:otsu_proxy}, we describe the procedure to determine the Otsu threshold for our case. Note that the tested thresholds $\cT$ correspond to the midpoints between consecutive loss values. Each of these thresholds divides the samples into two groups with at least two items each, which allows for computing the variance. Then, Otsu's method \cite{otsu_kmeans} exhaustively tests all thresholds and selects the one with a lower cost.

\subsection{Sample confidence}

We previously showed that $\ell_i^{\text{Proxy}}$ behaves as a bimodal distribution and that we can use Otsu's method \cite{otsu} to separate clean and noisy samples. Having this, we want to design a confidence score. Unlike SuperLoss \cite{superloss}, we advocate for computing a confidence score for each data point instead of doing so for each pair. Concretely, we want a confidence score $\sigma_i$ satisfying the following criteria:
\begin{enumerate}[label=(\roman*),font=\itshape]
    \itemsep0pt
    \item $\sigma_i$ is translation invariant \wrt $\ell_i^{\text{Proxy}}$. \label{c1}
    \item $\sigma_i \geq \sigma_j \Longleftrightarrow \ell_i^{\text{Proxy}} \leq \ell_j^{\text{Proxy}}$ ($i,j$ in the same batch). \label{c2}
    \item $\sigma_i \in [0, 1]$. \label{c3}
    \item As $\lambda \to 0$, $\sigma_i \to 1$ if clean, $\sigma_i \to 0$ otherwise. \label{c4}
    \item As $\lambda \to \infty$, $\sigma_i \to 1$. \label{c5}
\end{enumerate}

\begin{claim}
    The choice  
    \begin{align}
        \sigma_i := \exp\left\lbrace -W\left(  \left[\frac{\ell_i^{\text{Proxy}} - \tau}{2\lambda} \right]_+ \right) \right\rbrace\,,
        \label{eq:optimal_sigma_nca}
    \end{align}
    where $[\cdot]_+$ is the positive part, and $\tau$ computed with \cref{alg:otsu_proxy} satisfies \cref{c1,c2,c3,c4,c5}.
    \label{claim}
\end{claim}

\setlength{\textfloatsep}{5pt}%
\begin{algorithm}[t]
	\caption{\textsc{Computation of Otsu's threshold}}
	\label{alg:otsu_proxy}
	\begin{algorithmic}[1]
		\STATE \textbf{Inputs: }Proxy loss values $\{\ell_i^{\text{Proxy}}\}_i$
		\STATE \textbf{Output: }Threshold $\tau$
		\STATE Sort loss values $\texttt{L}\leftarrow\texttt{sorted}(\ell_i^{\text{Proxy}})$
		\STATE Define thresholds $\cT\leftarrow \{ \frac{\texttt{L[}i\texttt{]} + \texttt{L[}i+1\texttt{]}}{2} \}_{i\in \{2,3,\dots,|\cB| - 2\}}$
		\FORALL{$\tau' \in \cT$}
		    \STATE Let $\cC_0 \leftarrow \{\ell_i^{\text{Proxy}} | \ell_i^{\text{Proxy}} < \tau' \}$
		    \STATE Let $\cC_1 \leftarrow \{\ell_i^{\text{Proxy}} | \ell_i^{\text{Proxy}} \geq \tau' \}$
		    \STATE Let $\text{Cost}(\tau)\leftarrow \frac{1}{|\cB|}\left( |\cC_0| \text{Var}[\cC_0] + |\cC_1| \text{Var}[\cC_1] \right)$
		\ENDFOR
		\STATE $\tau \leftarrow \argmin_{\tau' \in \cT}\text{Cost}(\tau')$
	\end{algorithmic}
\end{algorithm}

\Cref{eq:optimal_sigma_nca} draws inspiration from SuperLoss \cite{superloss}. The reason is that the sample-level version of the SuperLoss confidence yields a clean expression and already satisfies \cref{c1,c2,c5}. While the proposed changes might seem subtle, they conceptually make a huge difference and improve the performance by a large margin (see \cref{tab:ablation}). Refer to the supplementary material for the proof of \cref{claim} and further discussion.

In stark contrast with SuperLoss \cite{superloss}, the confidence in ProcSim is not computed from the training loss. Having a different loss for the confidence computation and the parameter update can avoid biases, something considered in the works leveraging two models for unbiased noise sample identification \cite{han2018co,ibrahimi22,lee2017cleannet,wang2019,yu2019does,zeng2022}.

\begin{figure*}[t]
    \centering
    \includegraphics[width=.9\textwidth]{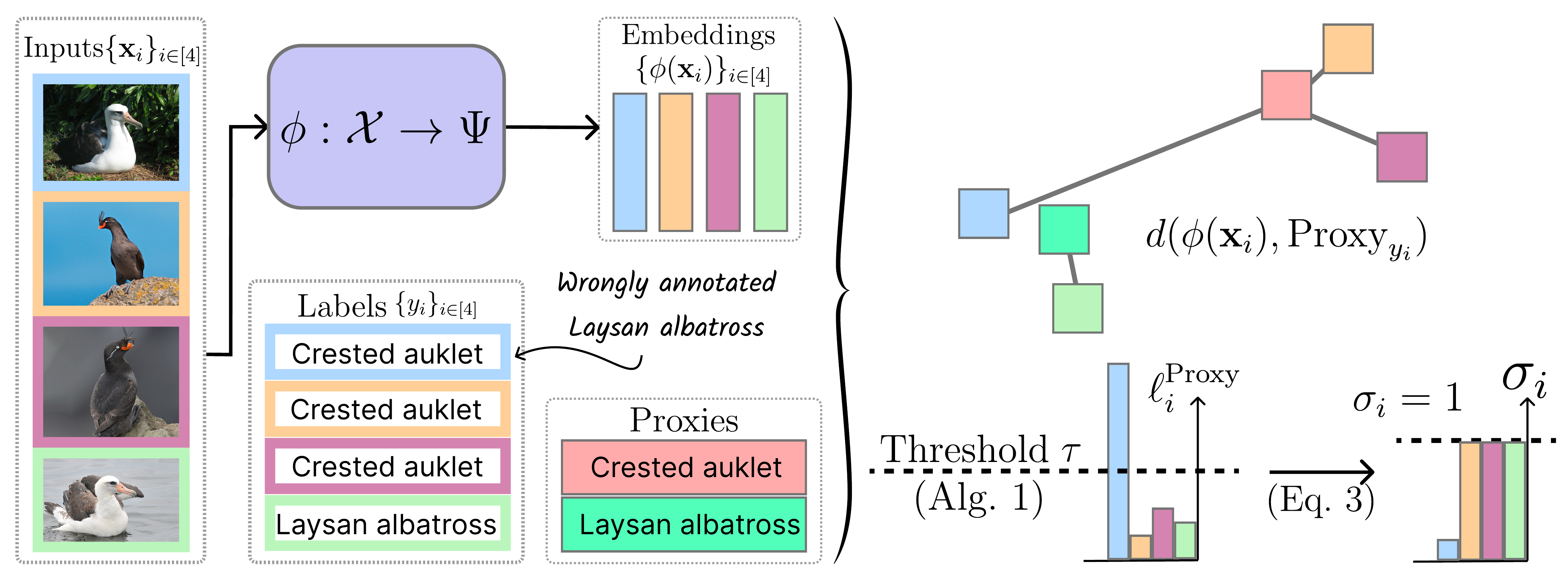}
    \vspace{-10pt}
    \caption{ProcSim model overview using an illustrative example. Here, we showcase the ProcSim model's functionality with four images $\{\xx_i\}_{i\in[4]}$ from the CUB200 dataset \cite{CUB_200_2011}. These images have class labels $y_1=y_2=y_3\neq y_4$, where $y_1$ has been erroneously assigned; it should be $y_1=y_4\neq y_2=y_3$. The \gls*{dml} model projects images into the metric space, yielding visual embeddings $\{\psi(\xx_i)\}_{i\in[4]}$. Then we compute the proxy loss $\ell_i^{\text{Proxy}}$, which is obtained by evaluating the distance from an embedding to its associated proxy. We determine a threshold for proxy loss values using \cref{alg:otsu_proxy}, and then calculate the sample confidence $\{\sigma_i\}_{i\in[4]}$ using \cref{eq:optimal_sigma_nca}. Samples with proxy loss values below the threshold possess unit confidence, while others have a smaller value that decreases as they move farther away from the proxies. Notably, $(\xx_1,y_1)$ is assigned a low confidence score, resulting in its limited contribution to updating the model parameters compared to other samples.}
    \label{fig:model_overview}
    \vspace{-10pt}
\end{figure*}

\subsection{ProcSim}
\label{sec:procsim}

ProcSim can work with any \gls*{dml} objectives writable as a sum over sample losses, a prerequisite for enabling independent scaling of the sample loss through $\sigma_i$. In this scenario, the gradients of the loss monotonously increase with $\sigma_i$, and low-confidence samples result in diminished gradient updates.

\gls*{dml} model training typically relies on binary similarities, \ie, identifying whether a pair of samples belong to the same class. However, evaluation involves unseen classes, so \gls*{dml} requires learning a notion of similarity rather than discriminating between training classes.

With this in mind, we add a self-supervised regularization loss to implicitly enforce a semantic structure among classes. Directly applying the confidence score to the regularized objective would alter the magnitude of both the supervised and unsupervised losses. Since the computation of the unsupervised loss does not rely on labels, we want it to be unaffected by the confidence.

The final objective then becomes
\begin{align}
    \cL = \frac{1}{|\cB|} \sum_{(\xx_i, y_i) \in \cB} \sigma_i \cdot \ell_i^{\text{DML}}  + \omega \ell_i^{\text{SSL}} \,,
    \label{eq:proposed_loss}
\end{align}
where $\omega$ a hyperparameter weighting the importance of the regularization loss. Note that setting $\sigma_i=1$ amounts to regular training, while for $\sigma_i=0$ the metric space is only learned with the semantic knowledge of the \gls*{llm}. An overview of the proposed method is presented in \cref{fig:model_overview}.

By default, ProcSim uses \gls*{ms} \cite{multisimilarity} as the supervised \gls*{dml} loss, but we also assess the performance using other losses in \cref{sec:ablation}. In the case of using the \gls*{ms} objective, the \gls*{dml} sample loss is
\begin{align}
	\begin{split}
		\ell_i^{\text{DML}} := &\frac{1}{\alpha}\log\left[ 1 + \sum_{j\in\cP_i}e^{-\alpha(S_{ij}-\delta)}\right] \\
		+ &\frac{1}{\beta}\log\left[ 1 + \sum_{j\in\cN_i}e^{-\beta(\delta-S_{ij})}\right]
	\end{split}\,,
	\label{eq:ms}
\end{align}
where $S_{ij}:=\lin{\phi(\xx_i), \phi(\xx_j)}$, which is equivalent to the cosine distance because we enforce $\norm{\phi(\xx_i)}=1 \ \forall i$, and $\alpha,\beta,\delta \in\R$ are hyperparameters.

Unless explicitly stated, we choose the \gls*{plg} loss \cite{language_guidance} as the self-supervised objective. To compute the \gls*{plg} loss images are input to a classifier pre-trained on ImageNet \cite{imagenet}. For each image, the top$-k$ class names are passed to the language part of CLIP \cite{clip} using the prompt ``\texttt{A photo of a \{label\}}".
Subsequently, $k$ similarity matrices are generated from the similarities of text embeddings. The \gls*{plg} loss is the row-wise KL divergence between the matrix of visual similarities and the mean of the $k$ matrices of language similarities. We refer the interested reader to the \gls*{plg} paper for further details.

\subsection{Semantically coherent noise generation}
\label{sec:semantic_noise_generation}

Artificial noise models allow injecting a controlled amount of noise to assess the robustness of different methods. A simple and ubiquitous noise model is the symmetric noise model \cite{symmetric_noise}, based on assigning an incorrect label picked uniformly at random from all the classes. However, labeling mistakes are often due to the semantic similarity of the correct and wrong classes. For this reason, noisy labels contained in real datasets follow a non-uniform distribution among classes, differing from the symmetric model. 

To mimic label errors where semantically similar images are confused, we propose computing the inherent taxonomy of the dataset's classes and using that in the noise injection process. Among the considered benchmark datasets, \gls*{sop} \cite{sop_dataset} is the only one that provides a grouping of classes. Concretely, the 22,634 products belong to one of twelve categories. Hence, for \gls*{sop}, one can inject semantic noise by swapping the class label of a training sample to another class in the train partition that falls within the same category.

To build a semantic taxonomy for the CUB200 \cite{CUB_200_2011} and Cars196 \cite{cars196} datasets, we group the natural language class names in the dataset by finding their hypernyms with WordNet \cite{wordnet}, as done by Rohrbach \etal \cite{wordnet_hypernoyms}. Given that a word can have multiple meanings, captured by WordNet synsets \cite{wordnet}, and hence several potential hypernyms, we ensure that all the class names are a hyponym of \textit{bird} and \textit{car}, respectively. In other words, we enforce a common root node grouping all the classes.
Refer to the supplementary material for further details and visualizations of the obtained class hierarchies.

To inject noise into the training splits of the datasets, we first filter the taxonomy to the training classes and treat each sample independently. Then, we traverse the class hierarchy starting at the leaf node corresponding to the original label until we find a node with several children. Finally, we select the incorrect class label uniformly at random among all children except the original class. We compute the class taxonomies only once and generate noisy versions of each dataset offline.

The fact that the noise model differs from the principles motivating ProcSim has two main reasons. On the one hand, using the same class hierarchy for noise generation and training could lead to unfair biases favoring our method. On the other hand, using word hierarchies such as WordNet \cite{wordnet} to resolve inter-class similarities empirically achieves lower retrieval performance than other methods such as \gls*{plg} \cite{language_guidance}.

\begin{table*}[t]
	\caption{Recall@1 on the CUB200 \cite{CUB_200_2011} dataset for different types and levels of noise. The methods included in the ablation study are classified depending on how the confidence (if any) is computed. All the methods in each group share the same hyperparameters. Best results are shown in \textbf{bold}. ProcSim and its variants consistently outperform all the other baselines, and ProcSim (base) achieves the best performance overall in terms of the harmonic mean on all corrupted datasets.}
	\vspace{-5pt}
	\centering
	\resizebox{.85\textwidth}{!}{
		\begin{tabular}{l || c || c | c | c || c | c | c || c}
             \toprule
             \multicolumn{1}{l}{\textsc{Noise type} $\rightarrow$} & \multicolumn{1}{l}{\textsc{None}} & \multicolumn{3}{c}{\textsc{Semantic}} & \multicolumn{3}{c}{\textsc{Uniform}} & \multicolumn{1}{|c}{\multirow{2}{5em}{\centering\textsc{Harmonic\newline mean}}} \\
             \cmidrule{1-8}
			\textsc{Methods} $\downarrow$ & - & 10\% & 20\% & 50\% & 10\% & 20\% & 50\% &  \\ 
			\midrule
			\hline
			\multicolumn{9}{>{\columncolor[gray]{.9}}l}{\textbf{Pair-level confidence}} \\
			\hline
			SuperLoss \cite{superloss} & 49.8 & 49.7 & 48.8 & 48.3 & 49.2 & 48.8 & 47.3 & 48.8 \\
			\midrule 
			\hline
			\multicolumn{9}{>{\columncolor[gray]{.9}}l}{\textbf{Base non-confidence-aware losses}} \\
			\hline
			Proxy-NCA \cite{proxynca} & 58.0 & 57.8 & 56.4 & 51.9 & 57.3 & 56.9 & 55.7 & 56.2\\
			\gls*{ms} \cite{multisimilarity} & 67.9 & 64.8 & 60.6 & 49.0 & 64.0 & 60.7 & 49.5 & 58.6 \\
			\gls*{ms} + \gls*{plg} \cite{language_guidance} & 69.4 & 68.7 & 67.7 & 62.3 & 68.5 & 68.4 & 55.5 & 65.4 \\
			\midrule
			\hline
			\multicolumn{9}{>{\columncolor[gray]{.9}}l}{\textbf{ProcSim and variants of it (ours)}} \\
			\hline
               ProcSim (base) & 70.1 & \textbf{72.2} & \textbf{71.0} & \textbf{67.9} & 69.3 & 70.4 & \textbf{60.8} & \textbf{68.6} \\
               \hline
               Threshold on \gls*{ms} instead of Proxy-NCA & 69.1 & 69.2 & 66.9 & 67.7 & 67.8 & 66.0 & 54.1 & 65.4 \\
               Proxy-NCA instead of \gls*{ms} as \gls*{dml} loss & 59.0 & 58.1 & 56.9 & 51.3 & 58.2 & 59.2 & 56.4 & 56.9 \\
			Regularization affected by confidence & 65.7 & 63.4 & 62.7 & 56.9 & 63.0 & 62.5 & 52.3 & 60.6 \\
                Global average instead of Otsu's method & 69.6 & 69.6 & 69.2 & 64.1 & \textbf{70.5} & 71.1 & 59.0 & 67.3 \\
                Gaussian Mixture Model instead of Otsu's method & \textbf{70.2} & 64.9 & 69.1 & 64.4 & 70.4 & \textbf{71.2} & 58.0 & 66.6 \\
            \bottomrule 
		\end{tabular}
	}
	\label{tab:ablation}
	\vspace{-10pt}
\end{table*}

\section{Experiments}
\label{sec:experiments}

\subsection{Experimental details}

\noindent\textbf{Datasets: }We report results on CUB200 \cite{CUB_200_2011}, Cars196 \cite{cars196}, and \gls*{sop} \cite{sop_dataset}. For all datasets, the sets of train and test classes are disjoint.

\noindent\textbf{Implementation details: }We implement ProcSim using PyTorch \cite{pytorch}, which also provides the utilized ResNet-50 \cite{resnet} backbone model with pre-trained ImageNet\cite{imagenet} weights. We replace the last layer of the backbone model with a fully connected layer that provides embeddings of dimension 512. The \gls*{plg} \cite{language_guidance} and the \gls*{ms} losses \cite{multisimilarity} are adapted from the original implementations and use the hyperparameters proposed by the authors for each dataset. The reported metrics are obtained by retrieving the nearest neighbors using the cosine similarity. For a fair comparison, we do not apply learning rate scheduling \cite{training_strategies}. We also report the results with fixed hyperparameters for each dataset to show that our method achieves good performance without requiring fine-tuning for different types and probabilities of noise. Please refer to the supplementary for additional implementation details.

\subsection{Ablation study}
\label{sec:ablation}

This section presents a study in which we assess the boost in image retrieval performance obtained with each of ProcSim's components. We report the Recall@1 achieved on the CUB200 \cite{CUB_200_2011} dataset and its corrupted versions in \cref{tab:ablation}. As baselines, we consider the base \gls*{dml} losses, which treat all samples equally, and SuperLoss \cite{superloss}. 

We implement the SuperLoss framework using the details provided by Castells \etal \cite{superloss}: learning rate, weight decay, scheduling\footnote{We apply learning rate scheduling to this method since its absence led to significantly worse results. All the other methods don't use scheduling to avoid confounders in the performance boost \cite{training_strategies}.}, contrastive loss \cite{chopra2005learning}, and $\lambda$ hyperparameter. We compute the contrastive loss using the PyTorch Metric Learning library \cite{musgrave2020pytorch} and weight each loss term by the confidence in \eqref{eq:optimal_sigma_pair} before reducing the loss.

SuperLoss \cite{superloss} yields poor results, which can be due to its susceptibility to techniques such as hard-negative mining and hyperparameter tuning \cite{ibrahimi22}. However, its surprising robustness to noise motivates the usage of a confidence-aware objective. Computing confidence scores at the sample level, as we do in ProcSim, yields much better results than the pair-level scheme of SuperLoss. Moreover, it can use any objective written as a sum over samples. Waiving this restriction allows the incorporation of more powerful \gls*{dml} objectives that alone outperform the pair-level confidence scheme.

\begin{table}[t]
	\caption{Recall@1 when ProcSim uses BERT \cite{bert} instead of CLIP \cite{clip} for the computation of the self-supervised loss. Difference with ProcSim inside parentheses.}
        \vspace{-10pt}
	\centering
	\resizebox{\linewidth}{!}{
		\begin{tabular}{l || c | c | c }
             \toprule
             \textsc{Uniform Noise (\%)} $\rightarrow$ & 10\% & 20\% & 50\% \\ 
			\midrule
			\hline
			\textsc{CUB200} \cite{CUB_200_2011} & 71.3 ({\color{green}+2.0}) & 71.2 ({\color{green}+0.8}) & 60.3 ({\color{red}-0.5}) \\
			\textsc{Cars196} \cite{cars196} & 86.9 ({\color{red}-0.3}) & 86.3 ({\color{green}+0.3}) & 75.6 ({\color{green}+0.4}) \\
			\textsc{SOP} \cite{sop_dataset} & 79.1 ({\color{red}-0.2}) & 77.9 ({\color{red}-0.5}) & 73.1 ({\color{red}-0.2}) \\
            \bottomrule
		\end{tabular}
	}
	\label{tab:bert}
\end{table}
\begin{table*}[t]
	\caption{Recall@$K$ (\%) on the benchmark datasets corrupted with 30\% uniform noise for different values of $K$. The reported results for all methods except ProcSim (ours) are taken from Yan \etal \cite{Yan2021AdaptiveHS}, and the asterisk ($^{*}$) indicates that their method was applied on top of the indicated \gls*{dml} loss. Best results are shown in \textbf{bold}. ProcSim achieves a superior performance according to most of the metrics. Note that, similarly to the runner-up method, ProcSim is a robustness framework built on top of the \gls*{ms} loss \cite{multisimilarity}.}
	\vspace{-5pt}
	\centering
	\resizebox{.9\textwidth}{!}{
	\begin{tabular}{l || c | c | c | c || c | c | c | c || c | c | c }
        \toprule
        \multicolumn{1}{l}{\textsc{Benchmarks} $\rightarrow$} & \multicolumn{4}{c}{\textsc{CUB200} \cite{CUB_200_2011}} & \multicolumn{4}{c}{\textsc{CARS196} \cite{cars196}} & \multicolumn{3}{c}{\textsc{\gls*{sop}} \cite{sop_dataset}} \\
        \midrule
		\textsc{Methods} $\downarrow$ & R@1  & R@2  & R@4 & R@8  & R@1  & R@2  & R@4  & R@8 & R@1  & R@10  & R@100  \\ 
		\midrule
		\hline
		Triplet	\cite{triplet_loss}&     54.3   &   67.1   &  77.4    &   85.6   &  44.3&   57.0   &   69.0   &   79.1   &  51.7        &  69.2    &  84.1     \\
		Triplet$^{*}$ \cite{Yan2021AdaptiveHS} & 55.5 & 68.1 & 78.2 & 85.9 & 46.1 & 58.2 & 69.6 & 79.3 & 52.9 & 70.1 & 84.6 \\
		\hline
		LiftedStruct \cite{sop_dataset}&    61.6   &  73.0    &  82.1    &   89.1   &  77.1   &  85.3    &   91.6   &  94.8    &     67.9      &   82.0   &  91.5  \\
		LiftedStruct$^{*}$ \cite{Yan2021AdaptiveHS} & 64.3 & 75.5 & 83.6 & 90.1 & 79.2 & 87.1 & 82.0 & 95.0 & 69.1 & 83.0 & 92.1 \\
		\hline
		MS 	\cite{multisimilarity}&   62.0 &    73.8  &   82.5   &   89.6     &  79.5    &  86.7    &  91.7    &   95.1   &  72.0   &   85.7     &   94.1 \\
		MS$^{*}$ \cite{Yan2021AdaptiveHS} &    65.3 &   76.1  & 84.7   &  90.7  & 82.4  &  89.5    &  93.8  & 95.9   &  73.6  &  86.9 &   94.8 \\\hline
		ProcSim (ours) & \textbf{68.8} & \textbf{79.8} & \textbf{87.4} & \textbf{92.4} & \textbf{84.1} & \textbf{90.6} & \textbf{94.7} & \textbf{97.0} & \textbf{77.7} & \textbf{89.5} & \textbf{95.0} \\
         \bottomrule
		\end{tabular}
	}
	\label{tab:yan}
	\vspace{-10pt}
\end{table*}
\begin{table*}[t]
    \centering
    \caption{Recall@1 (\%) on the benchmark datasets corrupted with different probabilities of uniform noise. The reported results for all methods except ProcSim (ours) are taken from the PRISM paper \cite{prism} and rounded to one decimal place for consistency with the other tables. Best results are shown in \textbf{bold}. While MCL+PRISM \cite{prism} performs slightly better than ProcSim for low levels of noise on \gls*{sop} \cite{sop_dataset}, our method consistently and considerably outperforms it in the other datasets.}
    \vspace{-5pt}
    \resizebox{.7\linewidth}{!}{ 
    \begin{tabular}{l || c | c | c || c | c | c || c | c | c }
    \toprule
    \multicolumn{1}{l}{\textsc{Benchmarks} $\rightarrow$} & \multicolumn{3}{c}{\textsc{CUB200} \cite{CUB_200_2011}} & \multicolumn{3}{c}{\textsc{CARS196} \cite{cars196}} & \multicolumn{3}{c}{\textsc{\gls*{sop}} \cite{sop_dataset}} \\
    \midrule
    \textsc{Methods} $\downarrow$ & 10\%           & 20\%           & 50\% & 10\%           & 20\%           & 50\% & 10\%           & 20\%           & 50\%           \\ \midrule
    \hline
    \multicolumn{10}{>{\columncolor[gray]{.9}}l}{\textbf{DML with Proxy-based Losses}} \\
    \hline
    FastAP \cite{cakir2019deep} & 54.1 & 53.7 & 51.2 & 66.7 & 66.4 & 58.9 & 69.2 & 67.9 & 65.8 \\
    nSoftmax \cite{zhaiclassification} & 52.0 & 49.7 & 42.8 & 72.7 & 70.1 & 54.8 & 70.1 & 68.9 & 57.3 \\
    ProxyNCA     \cite{proxynca}  & 47.1 & 46.6 & 41.6  & 69.8 & 70.3 & 61.8 & 71.1 & 69.5 & 61.5 \\
    Soft Triple    \cite{qian2019softtriple} & 51.9 & 49.1 & 41.5  & 76.2 & 71.8 & 52.5 & 68.6 & 55.2 & 38.5 \\
    \midrule
    \hline
    \multicolumn{10}{>{\columncolor[gray]{.9}}l}{\textbf{DML with Pair-based Losses}} \\
    \hline
    MS \cite{multisimilarity} & 57.4 & 54.5 & 40.7 & 66.3 & 67.1 & 38.2 & 69.9 & 67.6 & 59.6 \\
    Circle \cite{sun2020circle} & 47.5 & 45.3 & 13.0 & 71.0 & 56.2 & 15.2 & 72.8 & 70.5 & 41.2 \\
    Contrastive Loss  \cite{chopra2005learning}  & 51.8 & 51.5 & 38.6 & 72.3 & 70.9 & 22.9 & 68.7 & 68.8 & 61.2 \\
    MCL \cite{wang2020cross} & 56.7 & 50.7 & 31.2 & 74.2 & 69.2 & 46.9 & 79.0 & 76.6 & 67.2 \\
    MCL + PRISM \cite{prism} & 58.8 & 58.7 & 56.0 & 80.1 & 78.0 & 72.9 &\textbf{80.1} & \textbf{79.5} & 72.9 \\ 
    \hline
    ProcSim (ours) & \textbf{69.3} & \textbf{70.4} & \textbf{60.8} & \textbf{87.2} & \textbf{86.0} & \textbf{75.2} & 79.3 & 78.4 & \textbf{73.3} \\
    \bottomrule
    \end{tabular}
    }
    \label{tab:prism}
    \vspace{-8pt}
\end{table*}

Proxy-NCA loss \cite{proxynca} is preferable for noise identification, but its base performance falls behind the \gls*{ms} loss \cite{multisimilarity}. Adding the \gls*{plg} term \cite{language_guidance} promotes learning a representation that captures semantics. When using this regularization, we achieve a consistently better performance than plain \gls*{ms} loss and improved robustness against semantic noise compared to uniform noise.

We can see that weighting the \gls*{dml} loss by the confidence score and not on the regularization term yields a consistent improvement. In this case, noisy samples rely more on the regularization objective than the supervised \gls*{dml} loss, which is affected by label noise. Finally, using other thresholding methods like global average, as in SuperLoss \cite{superloss}, or Gaussian mixtures, as in \cite{li2020dividemix}, results in generally worse performance.

ProcSim does not have a monotonically decreasing performance with noise, a behavior only observed for the CUB200 dataset \cite{CUB_200_2011}. On the one hand, this can be due to using the same hyperparameters across all corrupted datasets and Otsu's method \cite{otsu} separating the samples into two groups. Note that this assumes that the Proxy-NCA loss \cite{proxynca} follows a bimodal distribution, which may decrease the contribution of correctly labeled samples when there are no wrong labels. Solving this is as easy as setting a larger $\lambda$, which accounts for a more equal treatment of the two sets of samples separated by the threshold. However, we wanted to show that even if not tuning $\lambda$, ProcSim obtains good results. Note that in any case finding $\lambda$ is equivalent to finding the noise level of the data, but to the severity by which we decrease the importance
of noisy sample. On the other hand, surprisingly, the best results are achieved with some semantic noise. Note that along with \gls*{plg} regularization, having some labels swapped to semantically similar samples can force the model to learn a space with semantically related groups.

\begin{table*}[t]
	\caption{Recall@1 (\%) on the benchmark datasets injected with different probabilities and models of noise. Best results are shown in \textbf{bold}. ProcSim obtains a consistently better performance and is significantly more robust to semantic noise than the alternatives.}
	\centering
	\vspace{-5pt}
	\resizebox{.7\linewidth}{!}{
		\begin{tabular}{l || c | c | c || c | c | c || c | c | c }
             \toprule
             \multicolumn{1}{l}{\textsc{Benchmarks} $\rightarrow$} & \multicolumn{3}{c}{\textsc{CUB200} \cite{CUB_200_2011}} & \multicolumn{3}{c}{\textsc{CARS196} \cite{cars196}} & \multicolumn{3}{c}{\textsc{\gls*{sop}} \cite{sop_dataset}} \\
             \midrule
			\textsc{Methods} $\downarrow$ & 10\% & 20\% & 50\% & 10\% & 20\% & 50\%  & 10\% & 20\% & 50\% \\ 
			\midrule
			\hline
			\multicolumn{10}{>{\columncolor[gray]{.9}}l}{\textbf{Uniform noise}} \\
            \hline
			LSD \cite{zeng2022} & 63.0 & 62.1 & 57.2 & 78.5 & 72.3 & 65.2 & 76.6 & 75.4 & 68.7 \\
			MCL + PRISM \cite{prism} & 58.1 & 56.4 & 54.7 & 78.7 & 74.8 & 68.6 & 76.4 & 76.6 & 72.6 \\
			\hline
			ProcSim (ours) & \textbf{69.3} & \textbf{70.4} & \textbf{60.8} & \textbf{87.2} & \textbf{86.0} & \textbf{75.2} & \textbf{79.3} & \textbf{78.4} & \textbf{73.3} \\
			\midrule
			\hline
			\multicolumn{10}{>{\columncolor[gray]{.9}}l}{\textbf{Semantic noise}} \\
            \hline
			LSD \cite{zeng2022} & 62.8 & 61.9 & 58.5 & 77.5 & 76.6 & 73.0 & 76.8 & 73.7 & 69.1 \\
			MCL + PRISM \cite{prism} & 57.7 & 57.9 & 50.6 & 77.8 & 75.9 & 63.4 & 76.6 & 75.8 & 72.2 \\
			\hline
			ProcSim (ours) & \textbf{72.2} & \textbf{71.0} & \textbf{67.9} & \textbf{86.9} & \textbf{86.3} & \textbf{81.1} & \textbf{79.0} & \textbf{77.8} & \textbf{73.3} \\
            \bottomrule
		\end{tabular}
	}
	\label{tab:semantic_vs_unif}
	\vspace{-15pt}
\end{table*}

\subsection{Influence of the language model}

The \gls*{plg} loss uses the language part of CLIP \cite{clip}, which is trained on vision-language paired datasets. While this means CLIP is well-aligned to learn semantic information for a visual similarity task, it also means that its training set might overlap with vision datasets \cite{clip}. For this reason, we tested ProcSim with a pre-trained BERT base model \cite{bert} as \gls*{llm}. The performance in \cref{tab:bert} shows the generalization capacity of ProcSim and factors out the possibility of unfair advantages by using CLIP.

Another possible issue arising from the \gls*{plg} loss is its limitation by the performance of the image classification model. Concretely, the classifier discretizes the number of language embeddings and limits it by the number of classes. Moreover, the categories may not align with the downstream dataset. One possible solution to bypass the classifier is to distill information from CLIP image embeddings. This approach takes advantage of the multi-modality of the model and achieves comparable performance in all datasets with slight improvements on \gls*{sop}. Refer to the supplementary for the results and additional discussions.

\subsection{Comparison to state-of-the-art}
\label{sec:sota}

Previous methods for robust \gls*{dml} report results on the benchmark datasets corrupted with uniform noise. For an extensive and exhaustive comparison, we present the image retrieval performance that ProcSim obtains compared to state-of-the-art approaches. We facsimile the results reported in the papers, which means that although the noise statistics are the same, the corrupted samples could differ. We also found methods like \gls*{ms} \cite{multisimilarity} to be inconsistent across papers, likely due to different implementations and hyperparameters.

\cref{tab:yan} presents the results obtained using adaptive hierarchical similarity \cite{Yan2021AdaptiveHS} on top of common \gls*{dml} objectives trained on datasets with a 30\% of wrong annotations. Among all \gls*{dml} objectives augmented with adaptive hierarchical similarity \cite{Yan2021AdaptiveHS}, \gls*{ms} attains the best performance, further motivating utilizing the \gls*{ms} loss as the base \gls*{dml} objective for ProcSim. The model trained with ProcSim outperforms all the other methods in all metrics, proving to be a better alternative to enhance the \gls*{ms} loss \cite{multisimilarity}.

Liu \etal \cite{prism} report results on the benchmark datasets corrupted with 10\%, 20\%, and 50\% of uniform noise. In \cref{tab:prism}, we report their results along the ProcSim performance. We can see further evidence of the superiority of \gls*{ms} \cite{multisimilarity} in front of Proxy-NCA loss \cite{proxynca} and of the vastly higher performance of Procsim on the CUB200 \cite{CUB_200_2011} and Cars196 \cite{cars196} datasets.

We can observe a slightly lower performance on \gls*{sop}. On the one hand, this is because the \gls*{sop} dataset is much more fine-grained than the others, and MCL + PRISM \cite{prism} is focusing on it and not on the other datasets, where ProcSim occasionally outperforms it by a 10\% difference. On the other hand, the \gls*{plg} is less effective on \gls*{sop} due to its higher class-to-sample ratio \cite{language_guidance}.

\subsection{Effect of semantic noise}

In \cref{tab:semantic_vs_unif}, we compare the effect of uniform and semantic noise on the state-of-the-art methods. To assess the performance of LSD \cite{zeng2022} and MCL + PRISM \cite{prism}, we use the code provided by the authors with the proposed hyperparameters and include the obtained results on uniform noise. MCL + PRISM \cite{prism} requires an estimate of the noise probability, and although not specified, we used the ground truth probabilities, thus favoring this method. Doing so achieved the closest results to those reported by Liu \etal \cite{prism} for CUB200 \cite{CUB_200_2011} and Cars196 \cite{cars196}, but not for \gls*{sop} \cite{sop_dataset}. We can observe that the results on the \gls*{sop} dataset \cite{sop_dataset} for both types of noise are alike as expected. The reason being that semantic noise assigns a label chosen uniformly at random over only one of the twelve categories for \gls*{sop}.

ProcSim attains the best performance in all cases. The competing approaches, especially MCL + PRISM \cite{prism}, are more affected by semantic noise. These results show that semantic noise can be more harmful as it generates samples with wrong labels that are harder to spot. Instead, ProcSim shows the opposite behavior, which we attribute to the resolution of inter-class relationships.

\section{Conclusions}
\label{sec:conclusions}

This paper proposed ProcSim, an approach for training \gls*{dml} models for visual search on datasets with wrong annotations. ProcSim is a confidence-aware framework that is usable on top of any \gls*{dml} loss to improve its performance on noisy datasets. ProcSim is superior to existing alternatives when applied to datasets with injected noise without even fine-tuning for different types and levels of noise.

This work also introduced a new noise model inspired by plausible labeling mistakes. The proposed semantic noise model yields samples with wrong class labels that are harder to spot and can occasionally be more harmful than the omnipresent uniform noise model. While real noise is complex and a mixture of different types of noise, including but not limited to semantic errors, we believe this is a step towards closing the gap between real-world and simulated noise.

\iftoggle{wacvfinal}{
\section*{Acknowledgments}
The authors thank Amit Kumar K C, Michael Huang, and René Vidal for fruitful discussions and useful suggestions.
O.B. is part of CLOTHILDE (``CLOTH manIpulation Learning from DEmonstrations") which has received funding from the European Research Council (ERC) under the European Union’s Horizon 2020 research and innovation program (Advanced Grant agreement No. 741930). O.B. thanks the European Laboratory for Learning and Intelligent Systems (ELLIS) for PhD program support.
}{}

{\small
\bibliographystyle{ieee_fullname}
\bibliography{bibliography}
}

\clearpage

\appendix

\section{Additional implementation details}
\label{sec:implementation_details}

This section provides further details on the implementation of ProcSim. We use the PyTorch framework \cite{pytorch} for all the components below.

\subsection{Data augmentation}
\label{sec:data_augmentation}
We perform standard data augmentation techniques as in previous \gls*{dml} works \cite{language_guidance, zeng2022,training_strategies}: random cropping to $224\times 224$ and horizontal flipping with probability 0.5.

\subsection{Model}

The \gls*{dml} model is a ResNet-50 \cite{resnet} in which we replaced the output classification layer with a fully connected layer that provides embeddings. The batch normalization layers have been frozen for improved convergence and stability across batch sizes \cite{training_strategies}. We take the ResNet-50 model implementation from the PyTorch library for computer vision \texttt{torchvision}, which also provides weights for ImageNet \cite{imagenet}. In particular, we use the second version of the pre-trained weights, \ie, \texttt{IMAGENET1K\_V2}. Throughout all the experiments, we use an embedding dimension of 512.

\subsection{Optimization}

We use the Adam \cite{adam} optimizer to update the parameters of the \gls*{dml} model. For CUB200 \cite{CUB_200_2011}, we train the model for 150 epochs with a base learning rate of $10^{-4}$. For both Cars196 \cite{cars196} and \gls*{sop} \cite{sop_dataset}, we use a base learning rate value of $10^{-5}$ and train for 250 epochs. In all cases, we use a weight decay \cite{weight_decay} of $4\cdot 10^{-4}$ and the default values in PyTorch \cite{pytorch} for the rest of the hyperparameters. We do not apply learning rate scheduling for unbiased comparison \cite{training_strategies}.

Proxies in Proxy-NCA are optimized independently using the Adam optimizer with all the default parameters. This choice is related to the observations Proxy-NCA++ \cite{teh2020proxynca++}, which indicate that using independent optimizers for updating the class proxies and the model parameters is one of the main drivers of performance that improves upon vanilla Proxy-NCA \cite{proxynca}.

The training process uses 4 NVIDIA Tesla V100 SXM2 16 GB GPUs with a batch size of 90 each. Note that the effective batch size is 360, which allows full utilization of the hardware at disposal for faster training and is typically not considered an influential factor of variation \cite{training_strategies}. While datasets with many classes like \gls*{sop} \cite{sop_dataset} may benefit from a larger batch size, Wang \etal \cite{multisimilarity} showed that when training a model with the \gls*{ms} loss, the performance on dataset like CUB200 \cite{CUB_200_2011} decreases with large batch sizes over 80.

\subsection{Loss}

The ProcSim loss is composed of two terms, as seen in \cref{eq:proposed_loss}. One such term is the supervised \gls*{dml} loss. By default, we use the \gls*{ms} loss \cite{multisimilarity}, \cf \cref{eq:ms}, with the hyperparameters proposed in the original paper: $\alpha=2$, $\beta=40$, and $\delta=0.1$. We adapt the original implementation\footnote{\url{https://github.com/msight-tech/research-ms-loss}} to perform batch operations and exclude pairs $(\xx_i,\xx_i)$ in $\cP$ instead of removing all pairs with a similarity higher than $1-\epsilon$, where we set $\epsilon=10^{-5}$.

The \gls*{plg} loss is computed using the original implementation\footnote{\url{https://github.com/ExplainableML/LanguageGuidance_for_DML}}, in which the language part of CLIP \cite{clip} (ViT-B/32 variant) is the chosen \gls*{llm}. In the experiment with the BERT language model in \cref{tab:bert}, we use the model and weights from hugging face\footnote{\url{https://huggingface.co/bert-base-uncased}}. The parameter $\omega$ scaling the \gls*{plg} loss is set to $\omega=10$ for CUB200 \cite{CUB_200_2011} and $\omega=5.5$ for Cars196 \cite{cars196}, the values reported in the official code repository. For \gls*{sop} \cite{sop_dataset}, they recommend using $\omega \in [0.1, 1]$, and we chose $\omega=0.5$ after testing with $\omega\in \{0.1,0.5,1.0\}$.

To compute the sample confidence, we treat $\tau$ and $\sigma$ as constant during backpropagation. We calculate the Proxy-NCA loss \cite{proxynca} using the PyTorch metric learning library \cite{musgrave2020pytorch} with the default hyperparameters. The value of $\lambda$ in \cref{eq:optimal_sigma_nca} determines how much the confidence of a sample decreases for losses greater than Otsu's threshold \cite{otsu}. Asymptotically, $\sigma_i \to 1$ as $\lambda \to \infty$, and as $\lambda \to 0$, $\sigma_i \to 0$ if $\ell_i^{\text{Proxy}} > \tau^{\text{Otsu}}$, and $\sigma_i \to 1$ if $\ell_i^{\text{Proxy}} \leq \tau^{\text{Otsu}}$. We tested $\lambda\in \{0.01,0.1,1.0,10.0\}$ and found the values of $\lambda=0.1$ on Cars196 \cite{cars196} and \gls*{sop} \cite{sop_dataset}, and $\lambda=1.0$ on CUB200 \cite{CUB_200_2011}, to give good performance across different levels of noise.

Note that a larger $\lambda$ on CUB200 \cite{CUB_200_2011}  implies that samples with a high loss are more penalized. This penalization explains the behavior observed in \cref{tab:ablation}, in which ProcSim obtained the best performance on noisy data. That is because the contribution of clean samples was potentially decreased in the absence of synthetic noise.

\section{Computation of confidence values}

\begin{proof}[Proof of Claim 1]
    \leavevmode
    The confidence score in \cref{eq:optimal_sigma_nca} is claimed to satisfy \cref{c1,c2,c3,c4,c5}. In the following, we prove each of these conditions:
    \begin{enumerate}[label=(\roman*),font=\itshape]
        \item $\sigma_i$ is translation invariant \wrt $\ell_i^{\text{Proxy}}$:\\[5pt]
        Note that each value $\ell_i^{\text{Proxy}}$ is subtracted by Otsu's threshold $\tau$. Thus, proving that $\tau$ is equivariant to translations of the proxy loss suffices (as those translations get canceled out). $\tau$ is computed as the cost minimizer threshold among those in $\cT$ (see \cref{alg:otsu_proxy}). $\cT$ are the midpoints between consecutive loss values. Hence, $\tau$ is translation equivariant. Finally, the cost is unaltered as the variance is translation invariant.
        \item $\sigma_i \geq \sigma_j \Longleftrightarrow \ell_i^{\text{Proxy}} \leq \ell_j^{\text{Proxy}}$ ($i,j$ in the same batch):\\[5pt]
        Since $(i,j)$ are in the same batch, they will share the same threshold $\tau$. Then, we have
        \begin{align}
            \frac{\ell_i^{\text{Proxy}} - \tau}{2\lambda} \leq \frac{\ell_j^{\text{Proxy}} - \tau}{2\lambda} &&\text{Since }\lambda\in\R_+\,.
        \end{align}
        The function $\max\left\lbrace0, \cdot\right\rbrace$ is increasing and hence the order is preserved. Its image is $\R_+$, and the restriction of $W(\cdot)$ to the domain of positive numbers is monotonously increasing. Therefore, for $a\leq b$
        \begin{align}
            W(a) \leq W(b) \Longleftrightarrow e^{-W(a)} \geq e^{-W(b)}\,,
        \end{align}
        since the exponential function is monotonously increasing.
        \item $\sigma_i \in [0, 1]$:\\[5pt]
        The image of the restriction of the Lambert W function to $\R_+$ is $[0, \infty)$, so the $\exp(\cdot)$ will be restricted to $(-\infty, 0]$. Therefore, $\sigma_i\in[0, 1]$ as claimed.
        \item As $\lambda \to 0$, $\sigma_i \to 1$ if clean, $\sigma_i \to 0$ otherwise:\\[5pt]
        The input of the Lambert W function
        \begin{align}
            \lim_{\lambda\to0^+}\frac{\ell_i^{\text{Proxy}} - \tau}{2\lambda} = \begin{cases}
            -\infty & \text{If } \ell_i^{\text{Proxy}} < \tau \\
            \infty & \text{Otherwise}
            \end{cases}\,,
        \end{align}
        where the first case corresponds to the definition of clean. Note that it cannot happen that $\ell_i^{\text{Proxy}} = \tau$ as the possible thresholds are mid-points between consecutive loss values. For the first case
        \begin{subequations}
        \begin{align}
            \lim_{x\to-\infty} &\exp\left[ -W\left( \max\left\lbrace 0, x \right\rbrace \right) \right] \label{eq:proof4_a} \\
            &= \exp\left[ -W\left( 0 \right) \right]  = \exp\left[ 0 \right] = 1\,,
            \label{eq:proof4_b}
        \end{align}
        \end{subequations}
        and for the second case
        \begin{subequations}
        \begin{align}
            \lim_{x\to\infty} &\exp\left[ -W\left( \max\left\lbrace 0, x \right\rbrace \right) \right] \\
            &=\lim_{x\to\infty}\exp\left[ -W\left(x \right) \right] \\
            &= \exp\left[ -\infty \right] = 0\,.
        \end{align}
        \end{subequations}
        \item As $\lambda \to \infty$, $\sigma_i \to 1$:\\[5pt]
        In this case, the input of the Lambert W function tends to 0, so we can leverage \cref{eq:proof4_a,eq:proof4_b}.
\end{enumerate}
\end{proof}

\begin{figure*}[t]
    \centering
    \begin{subfigure}{.5\textwidth}
      \centering
      \includegraphics[width=\textwidth]{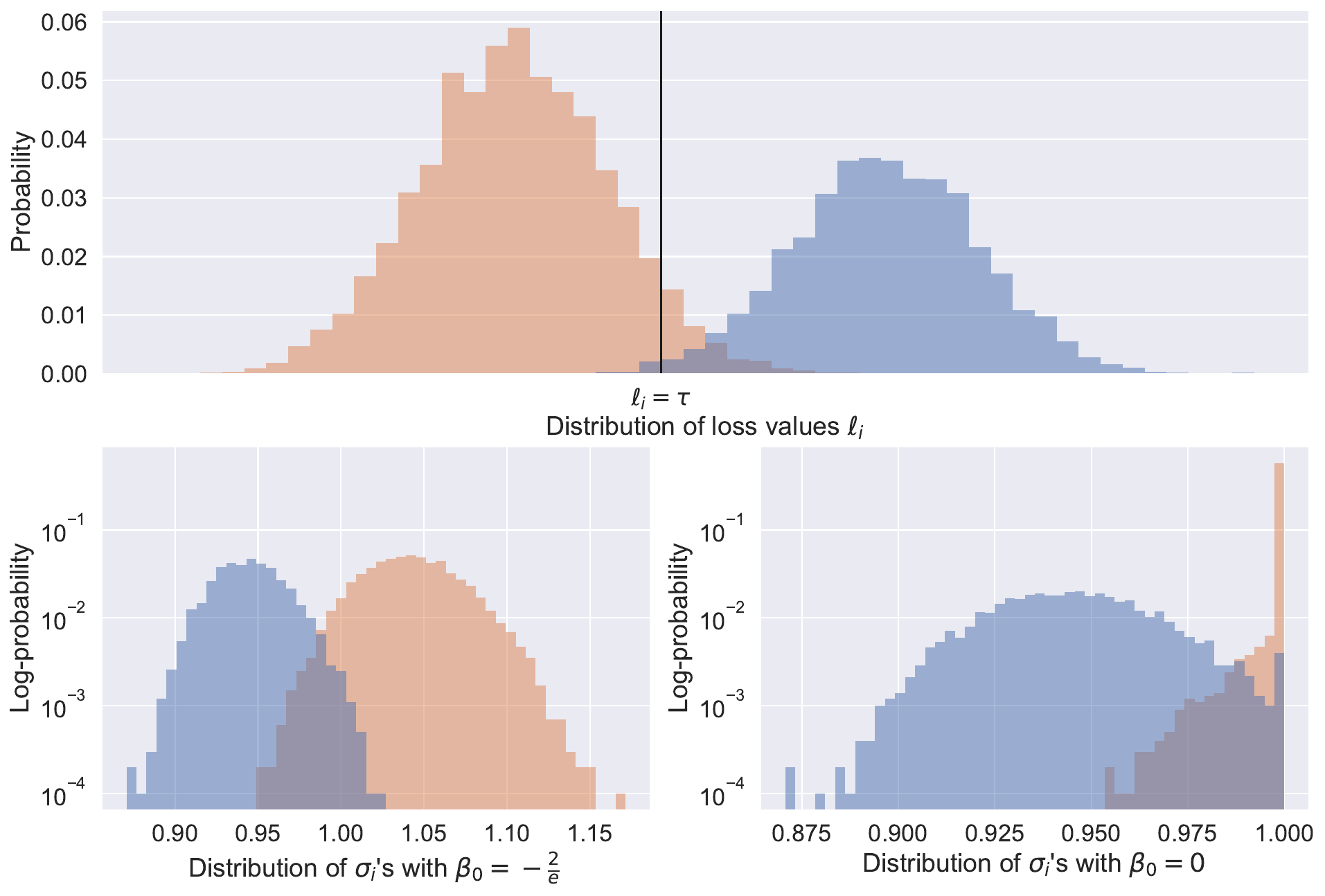}
      \caption{Non-separable distributions.}
      \label{fig:confidence_non-separable_example}
    \end{subfigure}%
    \begin{subfigure}{.5\textwidth}
      \centering
      \includegraphics[width=\textwidth]{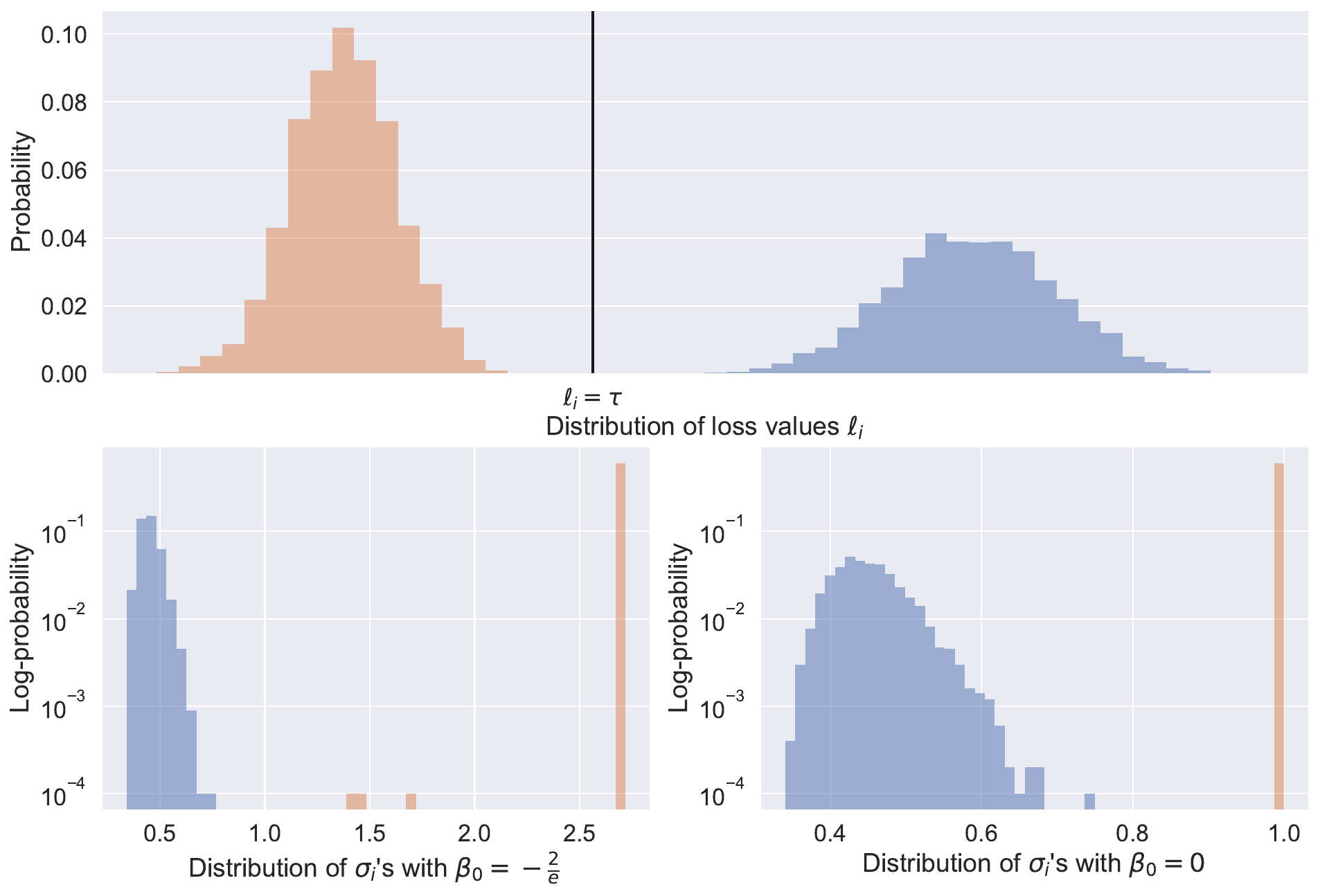}
      \caption{Separable distributions.}
      \label{fig:confidence_separable_example}
    \end{subfigure}
    \caption{Distribution of sample confidences computed using \cref{eq:optimal_sigma} with $\tau$ being the average of loss values and $\lambda=1$. In this toy example,
    the loss values follow a mixture of two Gaussians, shown in different colors. We had to decrease the precision of floating point numbers from 64 to 32 bits to avoid numerical errors for $\beta_0=-\frac{2}{e}$.}
    \label{fig:confidence_distribution_beta}
\end{figure*}

As acknowledged in the paper, the expression in \cref{eq:optimal_sigma_nca} is inspired by SuperLoss \cite{superloss}, which yields a simple and clean equation for the computation of the confidence that satisfies \cref{c1,c2,c5}. In the remainder of this section, we focus on the key differences between our confidence score and that of SuperLoss. While these changes might seem subtle, they conceptually make a huge difference and significantly improve performance (see \cref{tab:ablation}).

\subsection{Constraining the confidence}
\label{sec:superloss2}

As stated in \cref{c3}, we want to constrain the confidence $\sigma_i\in[0,1]$. Plugging the constraint into the sample-level confidence version of \cref{eq:superloss_objective} with constrained minimization, \ie
\begin{align}
\mathbb{E}_{i}\left[	\min_{\sigma_{i}\in \Sigma} (\ell_{i} - \tau_{i}) \sigma_{i} + \lambda (\log \sigma_{i})^2\right] \,,
\end{align}
yields an analytical expression to compute the confidence score corresponding to
\begin{align}
	\sigma_{i} = \exp\left[{-W\left(\frac{1}{2}\max\left\lbrace \beta_0, \frac{\ell_{ij} - {\tau_{ij}}}{\lambda}  \right\rbrace \right)}\right]\,,
	\label{eq:optimal_sigma}
\end{align}
where $\beta_0=-\frac{2}{e}$ when $\Sigma=\R$, as in SuperLoss \cite{superloss}, \cf \cref{eq:optimal_sigma_pair}. When $\Sigma=[0,1]$ as required by \cref{c3} we obtain $\beta_0=0$. By constraining the confidence, we avoid over-weighting the samples with a low loss and, at the same time, obtain the following desirable properties:

\paragraph{Asymptotic behavior: }With $\beta_0 = -\frac{2}{e}$ as in SuperLoss, as $\lambda \to 0$, $\sigma_i \to 0$ if $\ell_i > \tau$, $\sigma_i \to e$ if $\ell_i < \tau$, and $\sigma_i \to 1$ if $\ell_i=\tau$. Instead, with $\beta_0=0$, we satisfy \cref{c4}.

\paragraph{Numerical stability: }The evaluation of $W(\cdot)$ can become inaccurate close to $-\frac{1}{e}$, the so-called branch point. Particularly at the branch point, attained at $\ell_i - \tau \leq \lambda\beta_0$ with $\beta_0=-\frac{2}{e}$, the estimators used by well-known scientific computing libraries such as SciPy \cite{scipy} can fail to converge. The choice $\beta_0=0$ avoids these numerical problems.

\cref{fig:confidence_distribution_beta} presents a toy example illustrating the distributions with both values of $\beta_0$. When we have a bimodal distribution with separable modes (\cref{fig:confidence_separable_example}), selecting $\beta_0=0$ assigns a confidence of 1 to all samples with loss belonging to the distribution of a smaller mean. If the small loss assumption is satisfied, these loss values probably belong to clean samples, so we don't want to alter their contribution. The confidence score for the other samples can be controlled by $\lambda$ and be made arbitrarily close to 0.

In the non-separable case (\cref{fig:confidence_non-separable_example}), using $\beta_0=-2/e$ assigns diverse confidence scores to the samples belonging to the same distribution. By contrast, using $\beta_0=0$ assigns a unit confidence score to all values at the left of the threshold (the supposedly clean samples).

\begin{figure}[t]
    \centering
    \includegraphics[width=\columnwidth]{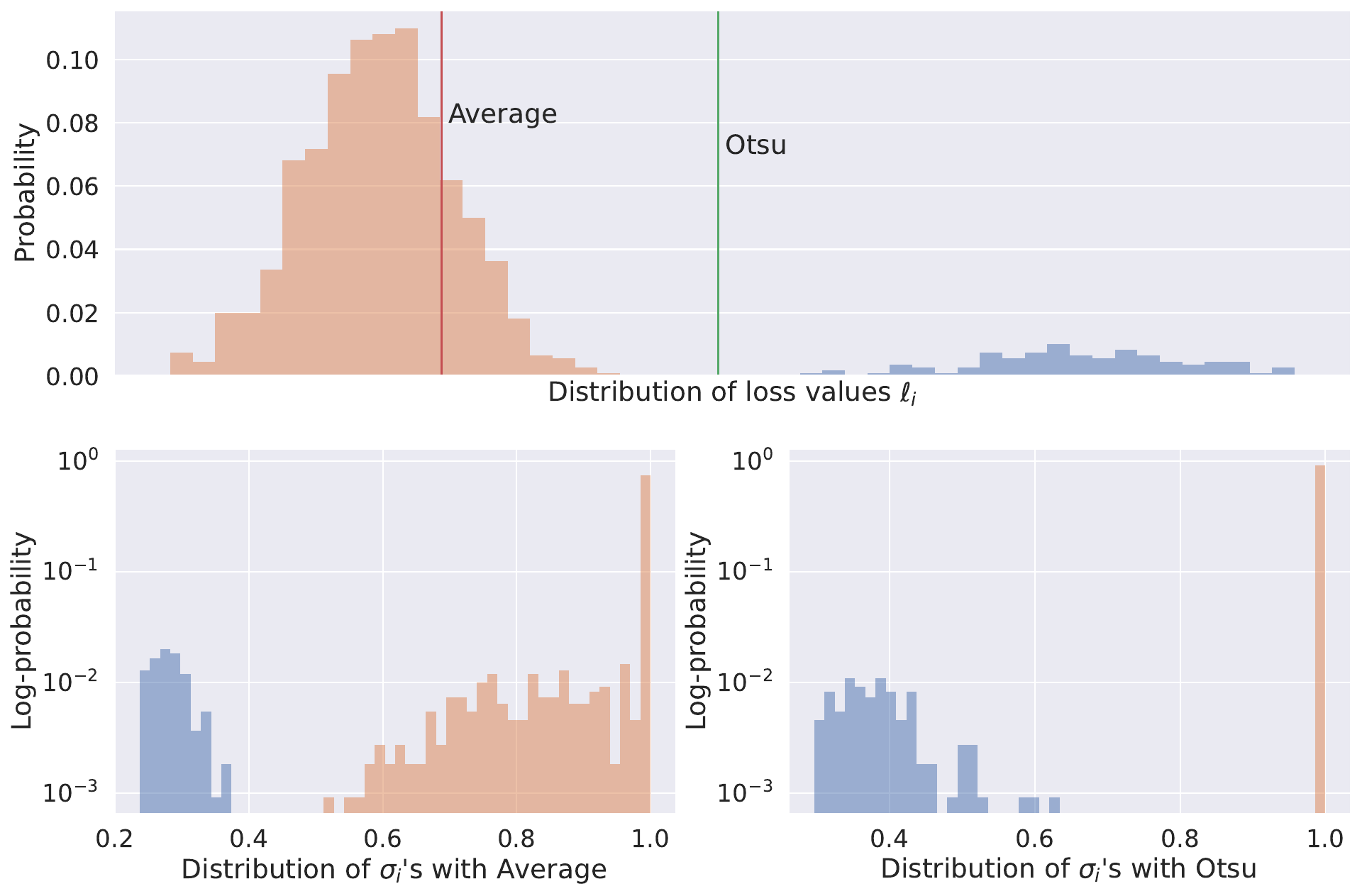}
    \caption{Distribution of sample confidences computed using \cref{eq:optimal_sigma} with $\tau$ being either the average as in SuperLoss \cite{superloss} or Otsu's threshold \cite{otsu} as in ProcSim. In both cases, we set $\beta_0=0$ and $\lambda=0.1$. In this toy example, the loss values follow a mixture of two Gaussians, shown in different colors.}
    \label{fig:confidence_distribution_threshold}
\end{figure}

\subsection{Thresholding}
\label{sec:superloss3}

Even if the loss can differentiate a wrong label and follows the ideal bimodal distribution, we can see that the global average is not suited. In \cref{fig:confidence_distribution_threshold}, we include a toy example to illustrate this observation, where we only consider one isolated iteration (so that the change of hard samples \wrt time is not an issue). Otsu's method selects $\tau$ based on the assumption that the distribution of losses is bimodal, which allows for treating clean and noisy samples differently.

Regarding the change of hard samples across iterations, we can also justify the choice of $\tau$ with a simple example. Imagine that the distribution of sample values is the same but just gets shifted. In the usual case, loss values decrease as training progresses, so the global average is larger than the average at a given iteration. Under this scenario, the number of samples whose contribution will be reduced decreases at every iteration. That is precisely the idea of curriculum learning, in which harder samples are included progressively at later training stages. However, it is not justifiable from the perspective of discerning clean from noisy samples since the number of noisy labels in a dataset stays constant.

\begin{figure}[t!]
    \centering
    \begin{subfigure}{\columnwidth}
      \centering
      \includegraphics[width=\columnwidth]{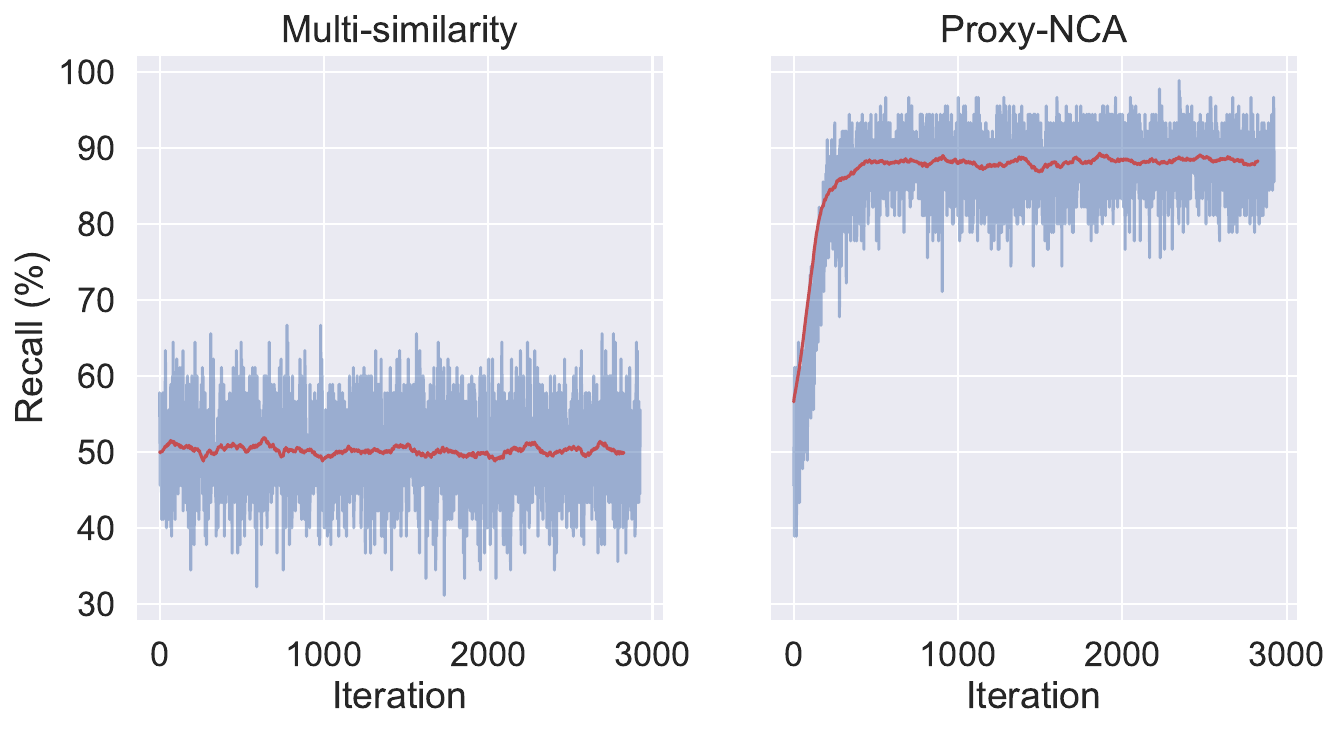}
      \caption{Results on CUB200 \cite{CUB_200_2011} with 50\% uniform noise.}
    \end{subfigure}%
    \\
    \begin{subfigure}{\columnwidth}
      \centering
      \includegraphics[width=\columnwidth]{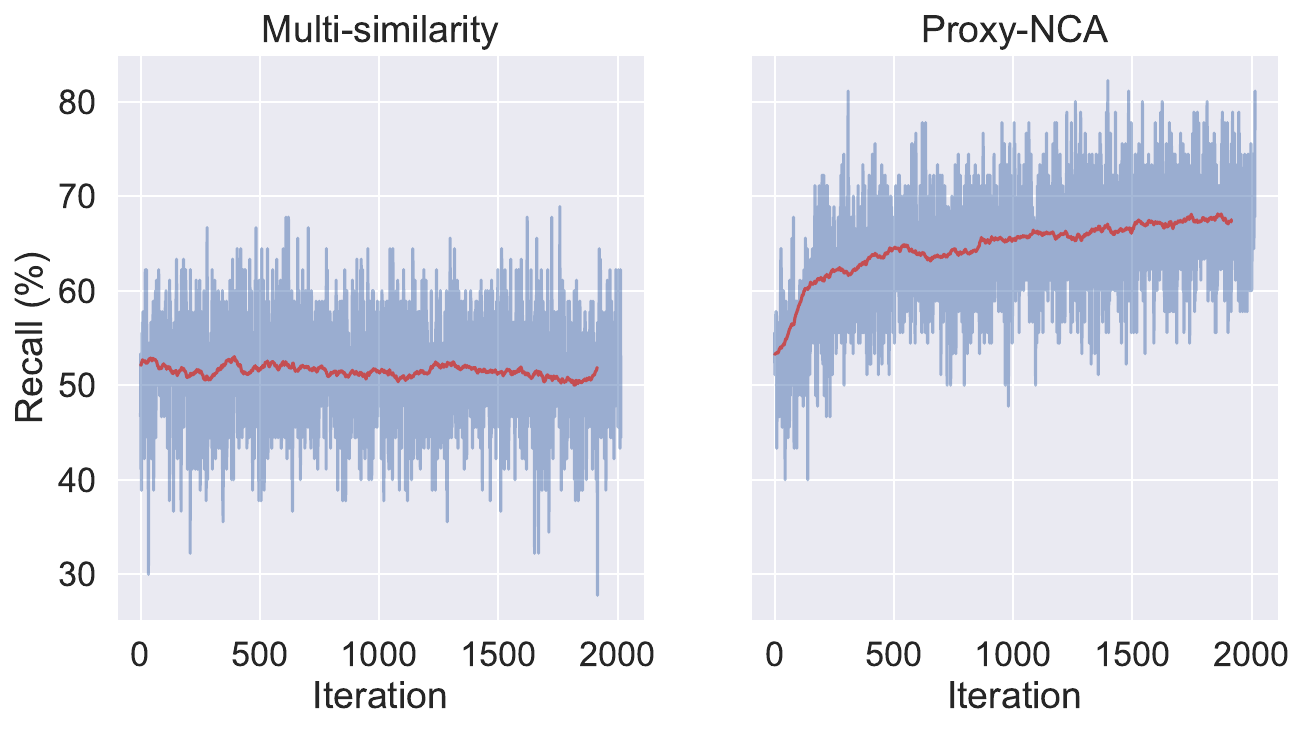}
      \caption{Results on CUB200 \cite{CUB_200_2011} with 50\% semantic noise.}
      \label{fig:otsu_hierarchical}
    \end{subfigure}
    \caption{Classification recall (\%) for the task of noisy sample identification using Otsu's method \cite{otsu}. The red line shows the moving average of the values obtained in a window of 100 iterations.}
    \label{fig:otsu_proxy}
\end{figure}

\subsection{Confidence score and training loss}
\label{sec:superloss4}

SuperLoss \cite{superloss} proposes minimizing $\ell_i \sigma_i$ treating the confidence score $\sigma_i$ as a constant and using any training objective $\ell_i$ for both the parameter update and the computation of $\sigma_i$. Consequently, if the training objective is composed of more than one term, they should be treated equally and as a whole. Instead, ProcSim applies different treatments to the supervised and self-supervised objectives implicated in the training loss. This simple modification is motivated by the fact that the self-supervised objective is unaffected by wrong annotations. Empirically this improves the \gls*{dml} performance, as seen in \cref{tab:ablation}.

Another notable difference with SuperLoss \cite{superloss} is that ProcSim disentangles the training loss and the objective used for the confidence computation. Doing so is similar to works relying on two independent models for unbiased noisy sample identification \cite{han2018co,ibrahimi22,lee2017cleannet,wang2019,yu2019does,zeng2022}. Moreover, it allows using losses with different properties.

On the one hand, we use Proxy-NCA loss \cite{proxynca} for its usefulness in noise identification, which is justified from a probabilistic perspective in \cref{sec:identify_noisy_samples} and empirically in \cref{fig:small_loss_trick}. For further evidence, even though ProcSim does not perform a hard classification into clean and noisy samples, we evaluated the usefulness of Otsu's method \cite{otsu} over Proxy-NCA \cite{proxynca} and \gls*{ms} \cite{multisimilarity} in the task of noisy sample identification. \cref{fig:otsu_proxy} depicts the evolution of the classification recall during training. As expected, we cab correctly identify most noisy samples by thresholding the Proxy-NCA loss \cite{proxynca}. However, using the same procedure on the \gls*{ms} loss \cite{multisimilarity} results in random classification. When the injected noise follows the semantic model proposed in this paper, \cref{fig:otsu_hierarchical} shows that Proxy-NCA is also better at spotting noisy samples, although the classification recall is significantly lower than when using the uniform noise model. We expected this behavior as semantic noise generates wrong labels that are harder to identify.

On the other hand, as shown in \cref{tab:ablation}, the base performance of Proxy-NCA \cite{proxynca} falls behind the \gls*{ms} loss \cite{multisimilarity}. At the same time, the \gls*{ms} loss is ineffective for spotting noisy samples, as shown in the example above. The ability to employ different and independent loss functions enhances the flexibility of ProcSim and enables us to leverage the strengths of various approaches, combining the best of both worlds.

\section{CLIP image embeddings}

\begin{table}[t]
	\caption{Recall@1 on the benchmark datasets for different levels of uniform noise when CLIP \cite{clip} image guidance is used instead of relying on an ImageNet classifier and an \gls*{llm}. Inside the parentheses, we indicate the performance difference with ProcSim.}
	\centering
	\resizebox{\linewidth}{!}{
		\begin{tabular}{l || c | c | c }
             \toprule
             \textsc{Noise level} $\rightarrow$ & 10\% & 20\% & 50\% \\ 
			\midrule
			\hline
			\textsc{CUB200} \cite{CUB_200_2011} & 67.5 ({\color{red}-1.8}) & 69.1 ({\color{red}-1.3}) & 60.5 ({\color{red}-0.3}) \\
			\textsc{Cars196} \cite{cars196} & 86.4 ({\color{red}-0.8}) & 85.5 ({\color{red}-0.5}) & 74.4 ({\color{red}-0.8}) \\
			\textsc{\gls*{sop}} \cite{sop_dataset} & 79.0 ({\color{red}-0.3}) & 77.9 ({\color{red}-0.5}) & 73.2 ({\color{red}-0.1}) \\
            \bottomrule
		\end{tabular}
	}
	\label{tab:clip}
\end{table}

The \gls*{plg} objective \cite{language_guidance} is a clever way to consider semantics to determine inter-class relationships. However, the number of ImageNet classes determines the maximum number of distinct language embeddings we can obtain with this procedure. ImageNet \cite{imagenet} covers a wide range of items but, especially when using datasets with low inter-class variations such as \gls*{sop} \cite{sop_dataset}, thousands of different classes fall into the same ImageNet category. The semantic ambiguity of those classes given by the language guidance regularization hinders resolving inter-class relations. In general, when the domain of the downstream task has little overlap with ImageNet classes, the resolution of inter-class relationships is somehow limited.

\begin{table*}[t]
    \centering
    \caption{Recall@1 (\%) on the benchmark datasets corrupted with different probabilities of uniform noise. The reported results for all methods except ProcSim (ours) are taken from the PRISM paper \cite{prism} and rounded to one decimal place for consistency with the other tables. Best results are shown in \textbf{bold}. While MCL+PRISM \cite{prism} performs slightly better than ProcSim for low levels of noise on \gls*{sop} \cite{sop_dataset}, our method consistently and considerably outperforms it in the other datasets.}
    \begin{tabular}{l || c | c | c || c | c | c || c | c | c }
    \toprule
    \multicolumn{1}{l}{\textsc{Benchmarks} $\rightarrow$} & \multicolumn{3}{c}{\textsc{CUB200} \cite{CUB_200_2011}} & \multicolumn{3}{c}{\textsc{CARS196} \cite{cars196}} & \multicolumn{3}{c}{\textsc{\gls*{sop}} \cite{sop_dataset}} \\
    \midrule
    \textsc{Methods} $\downarrow$ & 10\%           & 20\%           & 50\% & 10\%           & 20\%           & 50\% & 10\%           & 20\%           & 50\%           \\ \midrule
    \hline
    \multicolumn{10}{>{\columncolor[gray]{.9}}l}{\textbf{Algorithms for image classification under label noise}} \\
    \hline
    Co-teaching \cite{han2018co}  &  53.7  & 51.1  & 45.0 & 73.5  & 70.4  & 59.6  & 62.6  & 60.3  & 52.2  \\
    Co-teaching+  \cite{yu2019does}     & 53.3 & 51.0  & 45.2 & 71.5 & 69.6 & 62.4 &  63.4  & 67.9  & 58.3 \\
    Co-teaching \cite{han2018co} w/ Temperature \cite{zhaiclassification} & 55.6 & 54.2 & 50.7 & 77.5  & 76.3  & 66.9 & 73.7 & 72.0 & 64.1 \\   
    F-correction   \cite{patrini2017making} & 53.4 & 52.6 & 48.8 & 71.0 & 69.5 & 59.5 & 51.2 & 46.3 & 48.9 \\ 
    \midrule
    \hline
    \multicolumn{10}{>{\columncolor[gray]{.9}}l}{\textbf{DML with Proxy-based Losses}} \\
    \hline
    FastAP \cite{cakir2019deep} & 54.1 & 53.7 & 51.2 & 66.7 & 66.4 & 58.9 & 69.2 & 67.9 & 65.8 \\
    nSoftmax \cite{zhaiclassification} & 52.0 & 49.7 & 42.8 & 72.7 & 70.1 & 54.8 & 70.1 & 68.9 & 57.3 \\
    ProxyNCA     \cite{proxynca}  & 47.1 & 46.6 & 41.6  & 69.8 & 70.3 & 61.8 & 71.1 & 69.5 & 61.5 \\
    Soft Triple    \cite{qian2019softtriple} & 51.9 & 49.1 & 41.5  & 76.2 & 71.8 & 52.5 & 68.6 & 55.2 & 38.5 \\
    \midrule
    \hline
    \multicolumn{10}{>{\columncolor[gray]{.9}}l}{\textbf{DML with Pair-based Losses}} \\
    \hline
    MS \cite{multisimilarity} & 57.4 & 54.5 & 40.7 & 66.3 & 67.1 & 38.2 & 69.9 & 67.6 & 59.6 \\
    Circle \cite{sun2020circle} & 47.5 & 45.3 & 13.0 & 71.0 & 56.2 & 15.2 & 72.8 & 70.5 & 41.2 \\
    Contrastive Loss  \cite{chopra2005learning}  & 51.8 & 51.5 & 38.6 & 72.3 & 70.9 & 22.9 & 68.7 & 68.8 & 61.2 \\
    MCL \cite{wang2020cross} & 56.7 & 50.7 & 31.2 & 74.2 & 69.2 & 46.9 & 79.0 & 76.6 & 67.2 \\
    MCL + PRISM \cite{prism} & 58.8 & 58.7 & 56.0 & 80.1 & 78.0 & 72.9 &\textbf{80.1} & \textbf{79.5} & 72.9 \\ 
    \hline
    ProcSim (ours) & \textbf{69.3} & \textbf{70.4} & \textbf{60.8} & \textbf{87.2} & \textbf{86.0} & \textbf{75.2} & 79.3 & 78.4 & \textbf{73.3} \\
    \bottomrule
    \end{tabular}
    \label{tab:prism_extended}
\end{table*}
\begin{table*}[t]
    \caption{Performance of methods with ResNet-50 \cite{resnet} backbone and embedding dimension 512 on clean datasets. The best results are in \textbf{bold}. The results are taken from Roth \etal \cite{language_guidance}. Inside the parentheses, we indicate the boost in performance of ProcSim \wrt the mean performance of MS+\gls*{plg}, which is equivalent to setting unit confidence for all samples in the ProcSim framework (by letting $\lambda\to\infty$).}
    \centering
\resizebox{\textwidth}{!}{
\begin{tabular}{l || c | c | c || c | c | c || c | c | c}
     \toprule
     \multicolumn{1}{l}{\textsc{Benchmarks} $\rightarrow$} & \multicolumn{3}{c}{\textsc{CUB200} \cite{CUB_200_2011}} & \multicolumn{3}{c}{\textsc{CARS196} \cite{cars196}} & \multicolumn{3}{c}{\textsc{\gls*{sop}} \cite{sop_dataset}}\\
     \midrule
     \textsc{Methods} $\downarrow$ & R@1 & R@2 & NMI & R@1 & R@2 & NMI & R@1 & R@10 & NMI\\
     \midrule
     \hline
     EPSHN \cite{epshn}                     & 64.9 & 75.3 &  -   & 82.7 & 89.3 &  -  & 78.3 & 90.7 &  -    \\
     NormSoft \cite{zhaiclassification} & 61.3 & 73.9 &  -   & 84.2 & 90.4 &  -   & 78.2 & 90.6 &  -    \\
     DiVA \cite{milbich2020diva}            & 69.2 & 79.3 & 71.4 & 87.6 & 92.9 & 72.2 & 79.6 & 91.2 & 90.6 \\
     DCML-MDW \cite{Zheng_2021_CVPR_compositional} & 68.4 & 77.9 & 71.8 & 85.2 & 91.8 & 73.9 & 79.8 & 90.8 & 90.8 \\
     IB-DML \cite{seidenschwarz2021graphdml}
     & 70.3 & 80.3 & \textbf{74.0} & 88.1 & 93.3 & \textbf{74.8} & \textbf{81.4} & 91.3 & \textbf{92.6} \\
     \gls*{ms}+\textit{PLG} \cite{language_guidance} & 69.6 $\pm$  0.4 & 79.5 $\pm$ 0.2 & 70.7 $\pm$ 0.1 & 87.1 $\pm$ 0.2 & 92.3 $\pm$ 0.3 & 73.0 $\pm$ 0.2 & 79.0 $\pm$ 0.1 & 91.0 $\pm$ 0.1 & 90.0 $\pm$ 0.1\\     
     S2SD+\textit{PLG} \cite{language_guidance} & \textbf{71.4 $\pm$ 0.3} & \textbf{81.1 $\pm$ 0.2} & 73.5 $\pm$ 0.3 & \textbf{90.2 $\pm$ 0.3} & \textbf{94.4 $\pm$ 0.2} & 72.4 $\pm$ 0.3 & 81.3 $\pm$ 0.2 & \textbf{92.3 $\pm$ 0.2} & 91.1 $\pm$ 0.2\\  
     \hline
     ProcSim (ours) & 70.1 ({\color{green}+0.5})  & 79.6 ({\color{green}+0.1}) &  69.5 ({\color{red}-1.2}) & 87.7 ({\color{green}+0.6}) & 92.4 ({\color{green}+0.1}) & 72.2 ({\color{red}-0.8}) & 80.3 ({\color{green}+1.3}) & 91.4 ({\color{green}+0.4}) & 89.8 ({\color{red}-0.2}) \\
     \bottomrule
\end{tabular}
}
\label{tab:clean_table}
\end{table*}

ImageNet contains categories covering 2 or 3 classes in the CUB200 dataset \cite{CUB_200_2011}, such as \texttt{hummingbird}, \texttt{albatross}, \texttt{jay}, and \texttt{pelican}. We can observe a similar coverage for the Cars196 dataset \cite{cars196}, in which, e.g., \texttt{sports car}, \texttt{cab}, \texttt{wagon}, \texttt{convertible}, \texttt{land rover}, \texttt{racing car}, and \texttt{minivan} are present in ImageNet. This coverage provides a level of specificity that allows differentiating some of the classes and assessing their similarity. However, for the \gls*{sop} dataset \cite{sop_dataset}, we find superclasses such as \texttt{stapler} or \texttt{kettle} that, although being ImageNet categories, account for thousands of different classes. While some superclasses such as \texttt{chair}, \texttt{cabinet}, and \texttt{lamp} have multiple ImageNet classes adequate for each, the instance retrieval nature of \gls*{sop} and its large number of classes inside a superclass potentially reduces the knowledge transfer effectiveness.

\cref{tab:clip} presents the results obtained using CLIP image embeddings \cite{clip} instead of relying on a classifier and a language model. In this case, we bypass the ImageNet classifier and directly obtain embeddings encoding semantic information from images without limiting the number of different embeddings. We can see that this approach performs on par with standard \gls*{plg} on \gls*{sop} \cite{sop_dataset} but underperforms it on the other datasets.

\section{Additional comparisons}

In \cref{tab:prism}, we compared ProcSim to the methods reported in the PRISM paper \cite{prism}. For the sake of space, we excluded the algorithms for image classification under label noise. However, it may be interesting to compare these methods, especially those derived from Co-teaching \cite{han2018co}, which also relies on the small loss trick using the loss obtained by another model to have unbiased estimates. For this reason, we present all the results in \cref{tab:prism_extended}.

\begin{table*}[t]
	\caption{Recall@1 (\%) on the benchmark datasets corrupted with different types and probabilities of noise when Swin transformers \cite{swin} are used as backbone model. Best results shown in \textbf{bold}. Inside the parentheses, we indicate the boost in performance of ProcSim.}
	\centering
	\resizebox{\textwidth}{!}{
		\begin{tabular}{l || c || c | c | c || c | c | c | c }
             \toprule
             \multicolumn{1}{l}{\textsc{Noise type} $\rightarrow$} & \multicolumn{1}{l}{\textsc{None}} & \multicolumn{3}{c}{\textsc{Semantic}} & \multicolumn{3}{c}{\textsc{Uniform}} \\
             \midrule
			\textsc{Methods} $\downarrow$ & - & 10\% & 20\% & 50\% & 10\% & 20\% & 30\% & 50\%   \\ 
			\midrule
			\hline
			\multicolumn{9}{>{\columncolor[gray]{.9}}l}{\textbf{\textsc{CUB200} dataset \cite{CUB_200_2011}}} \\
            \hline
			\gls*{ms} \cite{multisimilarity} & 87.8 & 84.7 & 81.8 & 77.2 & 83.7 & 79.7 & 72.1 & 67.6 \\
			ProcSim (ours) & \textbf{88.4} ({\color{green}+0.6}) & \textbf{88.4} ({\color{green}+3.7}) & \textbf{88.5} ({\color{green}+6.7}) &\textbf{87.8} ({\color{green}+10.6}) & \textbf{88.1} ({\color{green}+4.4}) & \textbf{88.2} ({\color{green}+8.5}) & \textbf{87.1} ({\color{green}+15.0}) & \textbf{84.7} ({\color{green}+17.1}) \\
			\midrule
			\hline
			\multicolumn{9}{>{\columncolor[gray]{.9}}l}{\textbf{\textsc{Cars196} dataset \cite{cars196}}} \\
            \hline
			\gls*{ms} \cite{multisimilarity} & \textbf{92.0} & 88.9 & 85.0 & 71.5 & 88.9 & 83.3 & 78.1 & 46.7 \\
			ProcSim (ours) & 90.5 ({\color{red}-1.5}) & \textbf{89.3} ({\color{green}+0.4}) & \textbf{88.3} ({\color{green}+3.3}) & \textbf{85.1} ({\color{green}+13.6}) & \textbf{89.6} ({\color{green}+0.7}) & \textbf{87.5} ({\color{green}+4.2}) & \textbf{84.1} ({\color{green}+6.0}) & \textbf{69.7} ({\color{green}+23.0})\\
			\midrule
			\hline
			\multicolumn{9}{>{\columncolor[gray]{.9}}l}{\textbf{\textsc{SOP} dataset \cite{sop_dataset}}} \\
            \hline
			\gls*{ms} \cite{multisimilarity} & \textbf{84.3} & \textbf{83.5} & \textbf{82.6} & \textbf{77.7} & \textbf{83.4} & \textbf{82.3} & \textbf{81.3} & \textbf{78.3} \\
			ProcSim (ours) & 84.2 ({\color{red}-0.1}) & 82.3 ({\color{red}-1.2}) & 82.3 ({\color{red}-0.3}) & 77.6 ({\color{red}-0.1}) & 83.0 ({\color{red}-0.4}) & 82.1 ({\color{red}-0.2}) & 81.1 ({\color{red}-0.2}) & 77.5 ({\color{red}-0.8}) \\
            \bottomrule
		\end{tabular}
	}
	\label{tab:swin}
\end{table*}

ProcSim is a method for robust \gls*{dml} on noisy datasets. Nevertheless, for completeness, in \cref{tab:clean_table}, we include the obtained results on clean data side-by-side with state-of-the-art approaches. We present the methods with the same backbone architecture and embedding dimensionality as our current approach, as these are two of the main \gls*{dml}-independent drivers for generalization \cite{s2sd}. ProcSim offers comparable performance to state-of-the-art methods on clean data, although we focus on noisy datasets. In particular, ProcSim slightly improves the recall obtained without per-sample confidence, \ie, \gls*{ms}+\gls*{plg} \cite{language_guidance}.

Note that \gls*{nmi} slightly decreases when assigning confidence to samples. However, \gls*{nmi} varies across implementations and is sometimes uninformative \cite{musgrave2020metric}, so this metric has to be interpreted with caution.

The best method for clean data is S2SD+\gls*{plg}. S2SD \cite{s2sd} applies feature distillation between the output embeddings and embeddings computed with the so-called target networks, which results in higher-dimensional vectors. However, S2SD results in an objective expressed as a mean of losses for each target network. The fact that the mean is not over samples makes it incompatible with the presented framework.

\section{Usage with state-of-the-art backbone}

For a fair comparison, we performed all experiments using the standard ResNet-50 backbone \cite{resnet}. Nonetheless, when trying to get the best results, one can leverage more powerful and expressive backbones using modern architectures such as transformers \cite{transformers}. Swin transformers \cite{swin} are an example of these, and have been successfully applied to the visual retrieval task \cite{swintransfuse}.

\cref{tab:swin} shows the performance of ProcSim with Swin transformers \cite{swin} as the backbone model and the same hyperparameters used in the main paper for all the results with the ResNet-50 \cite{resnet} backbone (see details in \cref{sec:implementation_details}, where we specify the values for each of the three benchmark datasets). With no fine-tuning, ProcSim outperforms the base \gls*{ms} loss under the presence of noise for the CUB200 \cite{CUB_200_2011} and the Cars196 \cite{cars196} datasets. The difference in performance is monotonously increasing with the noise level and achieves an astounding increment of up to 23\% Recall@1 for the Cars196 \cite{cars196} dataset injected with 50\% uniform noise.

Consistently with the results obtained in the paper, the performance on the \gls*{sop} is somehow more limited. In this case, the base \gls*{ms} loss performs slightly better than ProcSim, although by at most 1.3\% of Recall@1. By selecting $\omega=0$ and $\lambda\to \infty$, ProcSim becomes \gls*{ms}. We could therefore match the performance of the plain \gls*{ms} loss and potentially obtain better results with some fine-tuning. However, we wanted to show the generalization capabilities of our method tailored only to each dataset regardless of the synthetic noise injected and the backbone.

\section{Obtaining class hierarchies}
\label{sec:hierarchies}

\begin{figure*}
    \centering
    \includegraphics[width=\textwidth]{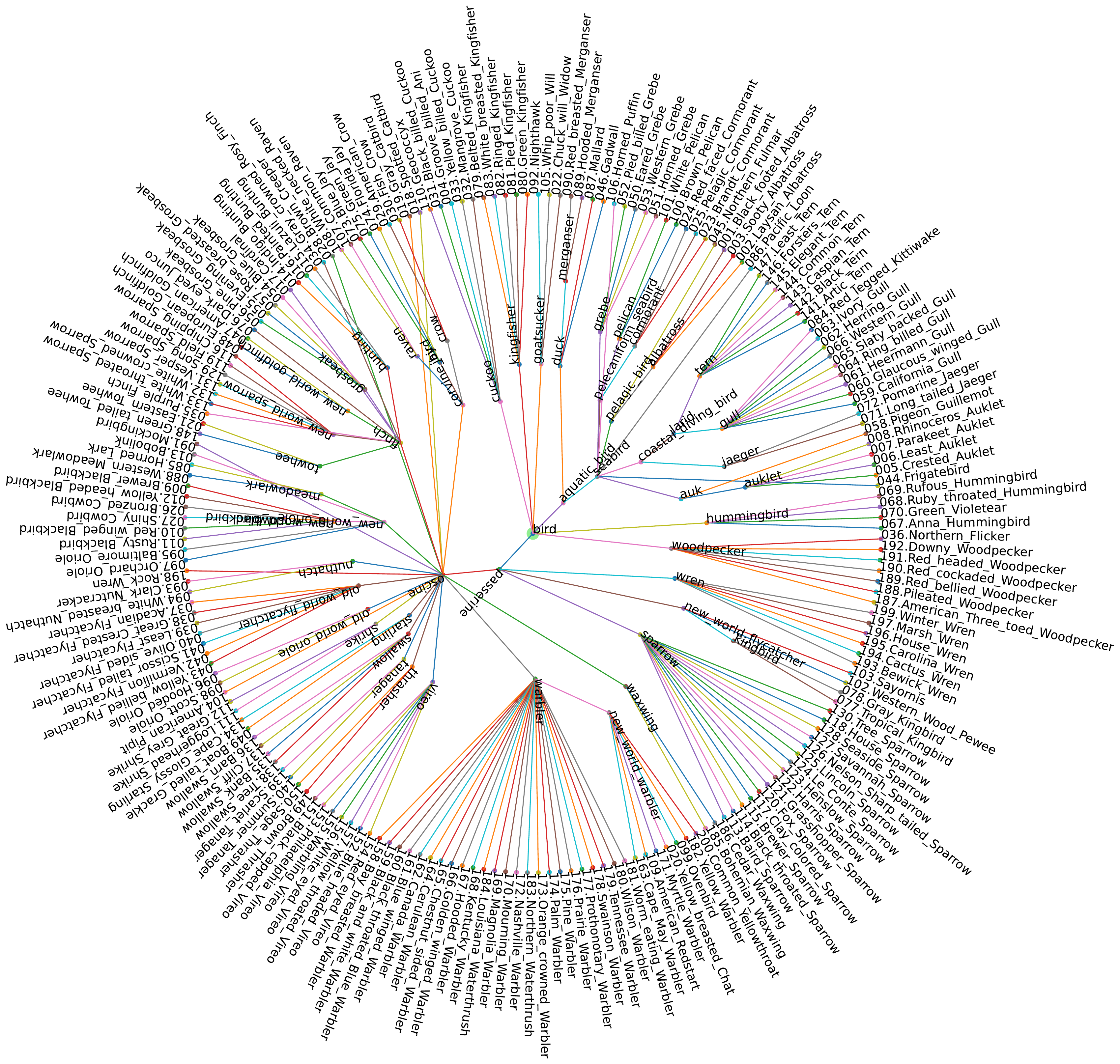}
    \caption{CUB200 \cite{CUB_200_2011} hierarchy.}
    \label{fig:cub_hierarchy}
\end{figure*}
\begin{figure*}
    \centering
    \includegraphics[width=\textwidth]{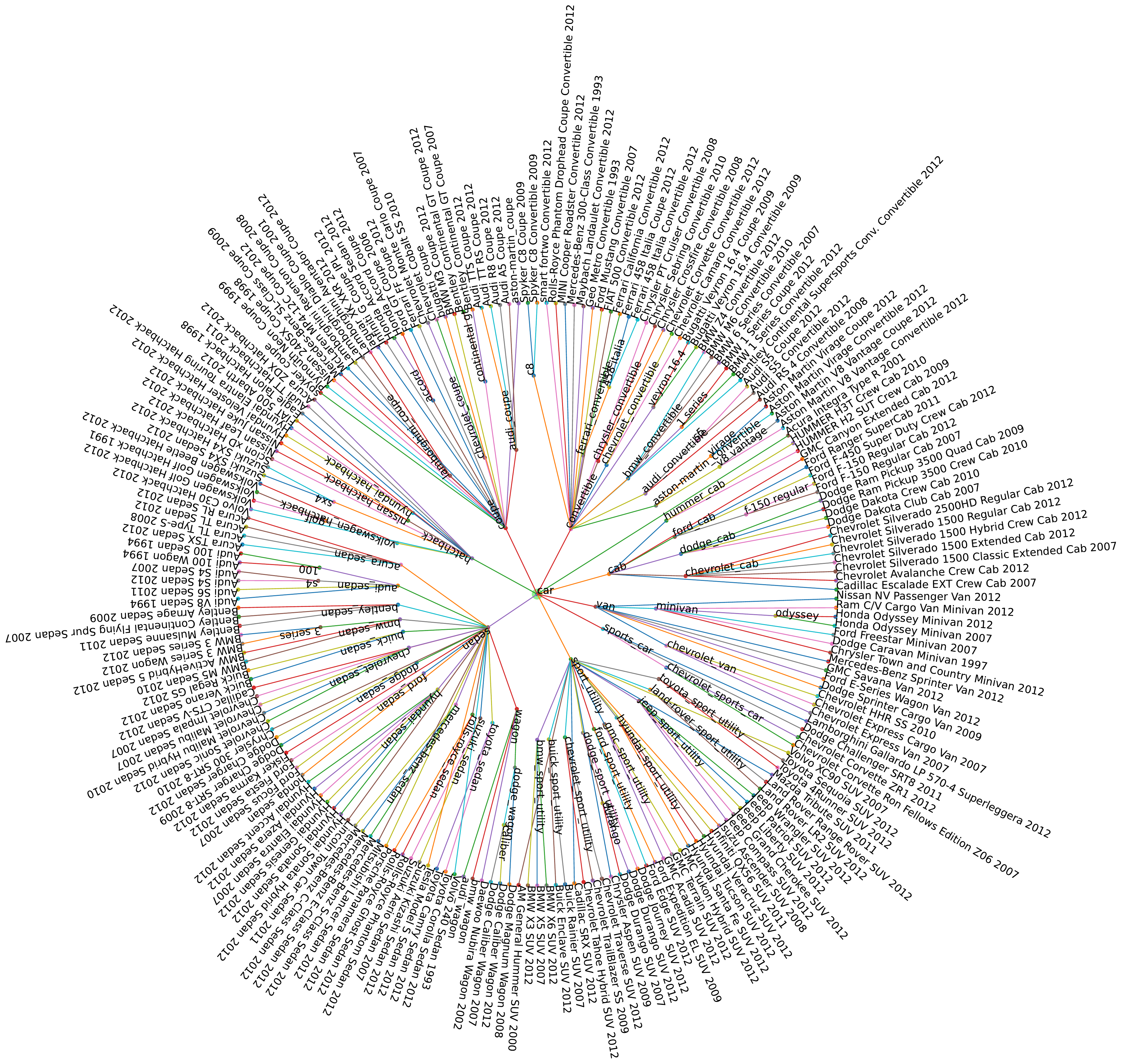}
    \caption{Cars196 \cite{cars196} hierarchy.}
    \label{fig:cars_hierarchy}
\end{figure*}

Finding class hierarchies is posed as a graph traversal problem and solved by depth-first search. Given natural language class names, we use WordNet to search all their possible meanings (with synsets) and semantic superclasses (with hypernyms). We consider each synset as a graph node and the hypernyms as oriented edges and keep exploring the graph according to the depth-first search algorithm. Among all possible paths in the graph resulting from different meanings of the class name or its superclasses, we select the one with a common hypernym across all dataset classes. Once we find this path, we stop looking for more possible synsets and hypernyms.

Note that class hierarchies are not used during training when applying ProcSim. They are needed only for the semantic noise model proposed in this paper, which aims at showing the robustness capabilities of ProcSim on benchmark datasets corrupted with more realistic noise.

Below, we provide details to obtain the hierarchy of classes for each dataset. We also show visualizations of the obtained class hierarchies. In them, we suppressed the nodes in the graph with a single child for better visualization.

\subsection{CUB200}
The CUB200 \cite{CUB_200_2011} dataset provides natural language class names consisting of bird types, and thus the common hypernym is \texttt{bird}. We first preprocess the class names to satisfy the expected input of WordNet \cite{wordnet}. Some classes are not included in WordNet \cite{wordnet}, in which case, we manually set the family of the species as a hypernym contained in the word corpora. \cref{fig:cub_hierarchy} depicts the CUB200 \cite{CUB_200_2011} hierarchy found using the described procedure.

\subsection{Cars196}
The Cars196 dataset \cite{cars196} contains classes whose common hypernym is \texttt{car} and have natural language names. However, the class labels contain other information like car type, brand, model, and year. Among all the class descriptors, only the car type is usable in WordNet \cite{wordnet}. Some brands may have different models of the same car type. Some models can also have different versions released over several years. With this in mind, we first group the classes by year, model, brand, and car type. Then, the car types are fed to WordNet \cite{wordnet} to find the complete class hierarchy, which we show in \cref{fig:cars_hierarchy}.

\subsection{SOP}
Unlike the other datasets, \gls*{sop} \cite{sop_dataset} does not contain natural language class names. Instead, the class names consist of numerical identifiers of the product. The only natural language description is in the form of categories. Since training and testing partitions have multiple classes for each category, we can inject semantic noise by only relying on those.

\end{document}